\documentclass{ieee-ojvt}
\usepackage[numbers,compress]{natbib}
\usepackage{amsmath,amssymb,amsfonts}
\usepackage{graphicx,color}
\usepackage[table]{xcolor}
\definecolor{lightgray}{gray}{0.9}
\definecolor{darkgray}{gray}{0.7}
\usepackage{arydshln}
\usepackage{hyperref}
\usepackage{booktabs}
\usepackage[caption=false,font=footnotesize]{subfig}
\usepackage[acronym,toc]{glossaries}

\glsdisablehyper

\newacronym{vlm}{VLM}{Vision-Language Model}
\newacronym{dtpqa}{DTPQA}{Distance-Annotated Traffic Perception Question Answering}
\newacronym{vqa}{VQA}{Visual Question Answering}
\newacronym{sota}{SOTA}{state-of-the-art}
\newacronym{llm}{LLM}{Large Language Model}
\newacronym{moe}{MoE}{Mixture of Experts}

\glsaddall

\usepackage{textcomp}
\def\BibTeX{{\rm B\kern-.05em{\sc i\kern-.025em b}\kern-.08em
    T\kern-.1667em\lower.7ex\hbox{E}\kern-.125emX}}
\AtBeginDocument{\definecolor{ojcolor}{cmyk}{0.93,0.59,0.15,0.02}}
\def\OJlogo{\vspace{-12pt}\includegraphics[height=24pt]{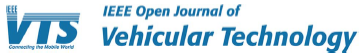}}
\definecolor{darkgreen}{rgb}{0,0.5,0}

\begin{document}
\receiveddate{XX Month, XXXX}
\reviseddate{XX Month, XXXX}
\accepteddate{XX Month, XXXX}
\publisheddate{XX Month, XXXX}
\currentdate{27 June, 2024}
\doiinfo{OJVT.2024.0627000}

\title{Evaluating Small Vision-Language Models on Distance-Dependent Traffic Perception}

\author{NIKOS~THEODORIDIS$^{1,2,3}$ (Graduate Student Member, IEEE), TIM~BROPHY$^{1,2,3}$ (Member, IEEE), REENU~MOHANDAS$^{1,2,3}$ (Member, IEEE),  GANESH~SISTU$^{1,2,4}$, FIACHRA~COLLINS$^{4}$, ANTHONY~SCANLAN$^{1,2}$, CIARÁN~EISING$^{1,2,3}$ (Senior Member, IEEE)}
\affil{Department of Electronic and Computer Engineering, University of Limerick, Castletroy, Co. Limerick V94 T9PX, Ireland}
\affil{Data Driven Computer Engineering Research Centre, University of Limerick, Castletroy, Co. Limerick V94 T9PX, Ireland}
\affil{Lero, The Irish Software Research Centre, University of Limerick, Limerick V94 NYD3, Ireland}
\affil{Valeo Vision Systems, Dunmore Road, Tuam, Co. Galway H54 Y276, Ireland}
\authornote{This work was supported in part by Science Foundation Ireland under Grant 13/RC/2094\_P2 and co-funded under the European Regional Development Fund
through the Southern and Eastern Regional Operational Programme to Lero - the Science Foundation Ireland Research Centre for Software, and in part by Valeo
Vision Systems.}
\markboth{EVALUATING SMALL VISION-LANGUAGE MODELS ON DISTANCE-DEPENDENT TRAFFIC PERCEPTION}{THEODORIDIS \textit{ET AL.}}

\corresp{CORRESPONDING AUTHOR: NIKOS THEODORIDIS (e-mail: theodoridis.nikolaos@ul.ie).}

\begin{abstract}
\glspl{vlm} are becoming increasingly powerful, demonstrating strong performance on a variety of tasks that require both visual and textual understanding. Their strong generalisation abilities make them a promising component for automated driving systems, which must handle unexpected corner cases. However, to be trusted in such safety-critical applications, a model must first possess a reliable perception system. Moreover, since critical objects and agents in traffic scenes are often at a distance, we require systems that are not “shortsighted”, i.e., systems with strong perception capabilities at both close (up to 20 meters) and long (30+ meters) range. With this in mind, we introduce \textbf{\gls{dtpqa}}, the first \gls{vqa} benchmark focused solely on perception-based questions in traffic scenes, enriched with distance annotations. By excluding questions that require reasoning, we ensure that model performance reflects perception capabilities alone. Since automated driving hardware has limited processing power and cannot support large \glspl{vlm}, our study centers on smaller \glspl{vlm}. More specifically, we evaluate several \gls{sota} small \glspl{vlm} on \gls{dtpqa} and show that, despite the simplicity of the questions, these models significantly underperform compared to humans (\(\sim \)60\% average accuracy for the best-performing small \gls{vlm} versus \(\sim \)85\% human performance). However, it is important to note that the human sample size was relatively small, which imposes statistical limitations. We also identify specific perception tasks, such as distinguishing left from right, that remain particularly challenging for these models. We hope our findings will encourage further research into improving the perception capabilities of small \glspl{vlm} in traffic scenarios, ultimately making them more suitable for automated driving applications.
\end{abstract}


\maketitle

\section{INTRODUCTION}
\IEEEPARstart{V}{ision}-Language Models have recently made remarkable progress across a wide range of tasks, from general \acrfull{vqa} to complex analysis and reasoning over scientific charts \cite{gpt4o2024, gemini2025, claude37sonnet2025, Wu2024c, Zhu2025d, Bai2025}. This strong performance is partly due to the pretrained \glspl{llm} that \acrfullpl{vlm} use as backbones, inheriting their powerful reasoning and generalisation abilities. The most common benchmarks used to evaluate \glspl{vlm} typically require a combination of strong perception and reasoning capabilities, as well as advanced world knowledge \cite{Liu2023d, Yue2024, Lu2024d, Yu2024b, Chen2024l, Liu2023e, Guan2024, Kembhavi2016}; however, this combination makes it difficult to pinpoint exactly where a model excels or falls short.

Some recent studies focus exclusively on the perception capabilities of \glspl{vlm} and show that these remain limited, particularly when it comes to fine-grained image details \cite{Rahmanzadehgervi2024, Tong2024, Gou2024, Kamoi2024, Kaduri2025}. The perception abilities of small \glspl{vlm} (fewer than 4 billion parameters) are even less explored, as most existing research focuses on large \acrfull{sota} models. This lack of attention hinders the adoption of \glspl{vlm}, especially smaller ones, in safety-critical applications that demand a robust and trustworthy perception system.

Automated driving is one such application where \glspl{vlm} could serve as a valuable component of the self-driving stack. The self-driving stack consists of the following layers \cite{Sharma2025}:

\begin{itemize}
    \item \textbf{Sensing}, which includes the sensors mounted on the car (cameras, LiDARs, radars, etc.).
    \item \textbf{Perception}, comprising algorithms (e.g., object detection, segmentation) that receive the sensory input and create a 3D map.
    \item \textbf{Localisation}, comprising algorithms that, based on the perception input, calculate the exact position and orientation of the ego-vehicle.
    \item \textbf{Planning}, comprising algorithms that, based on the perception input and knowledge of the ego-vehicle's position, plan its future trajectory.
    \item \textbf{Control}, which includes components that translate the defined plan into low-level actions (steering, accelerating, braking).
\end{itemize}

The strong reasoning and generalisation abilities of \glspl{vlm} make them a promising component for the planning layer in the self-driving stack \cite{Jiang2024c, Sima2024}. Specifically, a \gls{vlm} could be utilised as a high-level planner, where it first interprets the raw sensor input (i.e., perception) and then plans the desired action in natural language. This high-level decision could then be used, along with the raw sensory input, by other low-level components of the planning layer to predict the desired waypoints for the future trajectory. The work of Jiang \textit{et al.} investigates this exact use case \cite{Jiang2024c}.

However, the limited perceptual capabilities of current
\glspl{vlm} make them unreliable for deployment in high-risk
settings such as automated driving, even in expected scenarios. Even if a \gls{vlm} possesses strong reasoning abilities, it must first correctly perceive its environment before it can make appropriate driving decisions. For instance, consider a scenario that falls under the known-unsafe category of Safety Of The Intended Functionality (SOTIF): a pedestrian crossing the road (an expected scenario) at a long range (30-50 meters), where the model fails to detect them. In such a case, no matter how strong its reasoning capabilities are, they would be of little use in enabling the system to decide that it should brake.

Moreover, the hardware typically used in self-driving vehicles imposes strict constraints on the size of models that can be deployed on it. For instance, a commonly used hardware platform for self-driving cars is the NVIDIA Jetson Orin \cite{nvidiajetsonorin}. Even the most powerful variant, the Jetson AGX Orin, provides only 64 GB of unified memory shared between its CPU and GPU. \gls{sota} large \glspl{vlm} require significantly more memory. For example, InternVL3-78B \cite{Zhu2025d} would need approximately 156 GB of VRAM just to load its parameters in bfloat16 precision (2 bytes per parameter for 78 billion parameters), which far exceeds the memory capacity of the Jetson AGX Orin. When additional memory requirements, such as for activations storage and other components of the self-driving stack, are considered to be running concurrently, it becomes clear that deploying such large models is infeasible.

Due to these limitations, our study focuses exclusively on small \glspl{vlm} that can perform inference reliably on hardware like the Jetson AGX Orin and even on less powerful variants. More specifically, we study the perception-only capabilities of small, \gls{sota} \glspl{vlm} in traffic scenes. We assert that before trusting a \gls{vlm} to participate in crucial driving decisions, we should be confident that it has a strong perception system and can comprehend the complex visual input of a traffic scene. This also requires the model not to be ``shortsighted,'' as traffic scenes often include crucial objects and agents that are at a distance. We, therefore, ask the following question: \textbf{Can small \glspl{vlm} answer very simple (yet crucial) questions about traffic scenes that require only a good perception system, and how does their performance degrade as the distance of the object in question increases?} To investigate this, we need a \gls{vqa} benchmark that meets three specific criteria: (a) it must include trivial, perception-only visual questions that do not require reasoning; (b) all samples should relate to traffic scenes, with questions crucial for driving decisions (e.g., not questions like ``What colour is the car?''); and (c) it must contain distance annotations for each sample.

Since no existing \gls{vqa} benchmark satisfies all these requirements, we introduce a new benchmark: \acrfull{dtpqa}. Specifically, \gls{dtpqa} consists of trivial visual questions, most of which can be answered by any human at a glance, or after a closer look when there are distant objects, but are nonetheless crucial for driving decisions. Each \gls{dtpqa} sample consists of (a) an image, (b) a question, (c) the ground truth answer, and (d) the distance of the object in question, enabling analysis of how \gls{vlm} performance degrades with increasing object distance. 

Additionally, for these models to be trusted in safety-critical applications, their perception capabilities should not be sensitive to how a question is phrased. To this end, we investigate if small, semantically equivalent modifications to \gls{dtpqa} questions affect model performance on simple visual tasks, in order to assess their robustness.

To summarise, our contributions are as follows:
\begin{enumerate} 
    \item We introduce a novel benchmark, \gls{dtpqa}, which, to the best of our knowledge, is the first \gls{vqa} benchmark containing perception-only, trivial but crucial questions for traffic scenes with distance annotations. 
    \item We evaluate state-of-the-art small \glspl{vlm} on our benchmark and draw insightful conclusions from the results. 
    \item We measure human performance on our benchmark to establish its ceiling performance. 
    \item We examine if small, semantically preserving modifications to questions affect model performance on \gls{dtpqa}.
\end{enumerate}

\section{RELATED WORK}

VLMs have made remarkable progress in recent years by leveraging the power of strong \glspl{llm} as backbones. Most modern open-source \glspl{vlm} follow a similar high-level architecture that combines a visual encoder (typically a CLIP-like model \cite{Radford2021}) with an \gls{llm}, using a projector to map visual features into the input space of the \gls{llm} \cite{Liu2023b, Alayrac2022, Chen2024c, Bai2023, Lu2024c}. This approach has demonstrated impressive results across a variety of tasks, such as image captioning, visual question answering, visual reasoning, chart and diagram understanding, and document or Optical Character Recognition (OCR) based QA, that require integrating multiple skills. Simultaneously, the number of benchmarks for evaluating such models has grown significantly \cite{Liu2023d, Yue2024, Lu2024d, Yu2024b, Chen2024l, Liu2023e, Guan2024, Kembhavi2016}, most of which require a combination of perception, reasoning, and advanced world knowledge to be solved.

\subsection{Perception Capabilities of \glspl{vlm}}
A few studies have specifically investigated the perception capabilities of \glspl{vlm}. One line of research evaluates large \gls{sota} \glspl{vlm} on simple visual tasks. For example, Rahmanzadehgervi \textit{et al.} \cite{Rahmanzadehgervi2024} evaluated several \gls{sota} \glspl{vlm} on trivial visual questions, such as identifying the circled letter in a word or counting overlapping circles, that are easy for humans to answer at a glance. Their results showed that models like GPT-4o \cite{gpt4o2024} and Claude-3.5-Sonnet \cite{claude35sonnet2024} struggle significantly with such tasks. Kamoi \textit{et al.} \cite{Kamoi2024} introduced VisOnlyQA, a benchmark developed to assess the standalone visual perception abilities of \glspl{vlm}. They evaluated GPT-4o \cite{gpt4o2024}, Gemini 1.5 Pro \cite{gemini2024}, and InternVL2-76B \cite{internvl22024}, concluding that all of them exhibited weak perception capabilities. Moreover, they found that fine-tuning on perception-only benchmarks did not sufficiently address the problem. However, these studies focus exclusively on large \glspl{vlm} and not on traffic scenes, leaving a gap that our work aims to fill. In addition, none of these studies investigates how perception degrades with distance, which is one of the main focuses of our study.

Another line of research interprets the perception mechanisms of these models and explores ways to improve them. Kaduri \textit{et al.} \cite{Kaduri2025} analysed how \glspl{vlm} process visual information and found that generated tokens primarily attend to query (text) embeddings rather than image embeddings. As a result, the visual information is accessed indirectly through global context encoded in query embeddings (through their interaction with image embeddings) rather than directly from the image. Tong \textit{et al.} \cite{Tong2024} demonstrated that some image pairs are very close in CLIP’s \cite{Radford2021} embedding space but not in DINOv2’s \cite{Oquab2024}—a vision-only self-supervised model. They showed that such image pairs often lead \glspl{vlm} to make mistakes and proposed a new visual encoding strategy that incorporates both CLIP and DINOv2 features. Tong \textit{et al.} \cite{Tong2024b} conducted a thorough study on what contributes to better visual representations by examining different vision encoders, connectors, training datasets, and training recipes. They also introduced a new type of connector that combines features from multiple vision encoders while incorporating an inductive spatial bias, and showed that this improves the resulting visual representations. Finally, Chen \textit{et al.} \cite{Chen2025c} investigated why spatial reasoning is challenging for \glspl{vlm} by studying the models’ attention maps. They demonstrated that smoothing the attention when the model is uncertain in its prediction and sharpening it otherwise is an effective training-free method for improving \gls{vlm} performance on tasks that require spatial reasoning. However, these studies also focus on large models and do not include models with fewer than 4B parameters in their experiments.

\subsection{VLMs in Automated Driving}
Numerous studies have explored the use of \glspl{vlm} in automated driving \cite{Jiang2024c, Sima2024, Jin2023b, Zheng2024, Jiao2024b, Wang2024h, Zhou2025, Marcu2024, gq7xi4, Qiao2025, Xie2025, Tom2023, Chen2025, Guo2024, Ding2024, Charoenpitaks2025, Choudhary2023, Gopalkrishnan2024, Lbberstedt2025, Zhou2025b}. The most straightforward approach involves the use of \glspl{vlm} in end-to-end driving, where it takes a traffic scene as input and outputs low-level actions. For instance, Sima \textit{et al.} \cite{Sima2024} decomposed the driving task into several stages, such as perception, high-level planning, and waypoint prediction, and trained a \gls{vlm} to perform each step, enabling it to ``control'' a vehicle. Similarly, Jin \textit{et al.} \cite{Jin2023b} trained a \gls{vlm} to predict driving actions and generate textual descriptions and justifications for them. A slightly different approach was taken by Jiang \textit{et al.} \cite{Jiang2024c}, who used a \gls{vlm} for high-level planning in natural language. They then encoded this high-level plan and fed it, along with sensory input, into a trajectory prediction module. Their results showed that this method outperforms using only the sensory input.

Other studies have similarly investigated the application of \glspl{vlm} to automated driving. Notable examples include: Choudhary \textit{et al.} \cite{Choudhary2023} combined features extracted from VLMs with Bird’s-Eye View (BEV) maps and used these augmented BEV maps as input to an LLM to analyse the visual scene. They evaluated its performance on perception, visual reasoning, and decision-making tasks. Gopalkrishnan \textit{et al.} \cite{Gopalkrishnan2024} introduced EM-VLM4AD, a lightweight \gls{vlm}, and trained and evaluated it on a variety of tasks, including perception, planning, and prediction. Lübberstedt \textit{et al.} \cite{Lbberstedt2025} introduced V3LMA, an approach that combines \glspl{llm} and \glspl{vlm} to enhance the 3D scene understanding of AI systems. This is achieved by leveraging text descriptions from object detections and video inputs alongside visual features from \glspl{vlm}. Finally, Zhou \textit{et al.} \cite{Zhou2025b} addressed the problem of fine-grained information loss in the vision encoder due to downsampling while ensuring computational tractability. They proposed a novel architecture that uses a dynamic-resolution image input processing module. All these studies train and evaluate models across a variety of tasks without focusing solely on perception capabilities.

Another line of research focuses on developing suitable benchmarks for assessing and evaluating \glspl{vlm} in traffic scenarios. Marcu \textit{et al.} \cite{Marcu2024} introduced LingoQA, a \gls{vqa} benchmark for automated driving, and showed that \gls{sota} \glspl{vlm} still lag behind human performance. Xie \textit{et al.} \cite{Xie2025} proposed DriveBench, a benchmark for evaluating skills such as perception, prediction, and planning, and found that \glspl{vlm} often produce plausible answers based on world knowledge rather than grounded visual input. Tom \textit{et al.} \cite{Tom2023} developed RoadTextVQA, a Video-QA benchmark focusing on reading road text and recognising traffic signs, and their evaluations revealed significant room for improvement. Chen \textit{et al.} \cite{Chen2025} introduced CODA-LM, a benchmark for assessing \glspl{vlm} on self-driving corner cases, and tested various \gls{sota} models. Guo \textit{et al.} \cite{Guo2024} presented DriveLLM, targeting spatial understanding in traffic scenes. Ding \textit{et al.} \cite{Ding2024} proposed NuInstruct, a challenging benchmark requiring temporal, multiview, and spatial reasoning, and introduced a method for enhancing \glspl{vlm} by integrating Bird’s-Eye View (BEV) features. Finally, Charoenpitaks \textit{et al.} \cite{Charoenpitaks2025} introduced TB-Bench, a benchmark for evaluating \glspl{vlm}' spatiotemporal understanding in traffic scenes, and their results highlighted poor performance among \gls{sota} models. Most of these works focus on a holistic evaluation of the skills required for autonomous driving. However, this comprehensive approach makes it difficult to pinpoint specific weaknesses, as all skills are simultaneously required to solve a task. In contrast, DTPQA isolates and evaluates the perception capabilities of models in very simple visual tasks. Furthermore, DTPQA facilitates the evaluation of how these models' perception capabilities degrade with increasing distance, an aspect crucial for traffic scenarios and, to the best of our knowledge, absent in the existing literature.

\section{DTPQA}
\label{section_dtpqa}
\glsentrylong{dtpqa} is a \gls{vqa} benchmark designed to evaluate the perception-only capabilities of \glspl{vlm} in traffic scenarios, using trivial yet crucial questions relevant to driving decisions. It consists of two parts: a synthetic benchmark (DTP-Synthetic) created using the CARLA simulator \cite{Dosovitskiy2017}, and a real-world benchmark (DTP-Real) built on top of nuScenes \cite{Caesar2020}. Both parts are single-image \gls{vqa} benchmarks aimed at assessing the perception capabilities of \glspl{vlm} without requiring any reasoning skills.

A key feature of \gls{dtpqa} is that it includes variations of the same question across different object distances, ranging from 5 to 50 meters, with intermediate levels at 10, 20, 30, and 40 meters. It also includes cases where the object is completely absent, referred to as \textit{negative samples}.

\begin{figure}              \centerline{\includegraphics[width=3.5in]{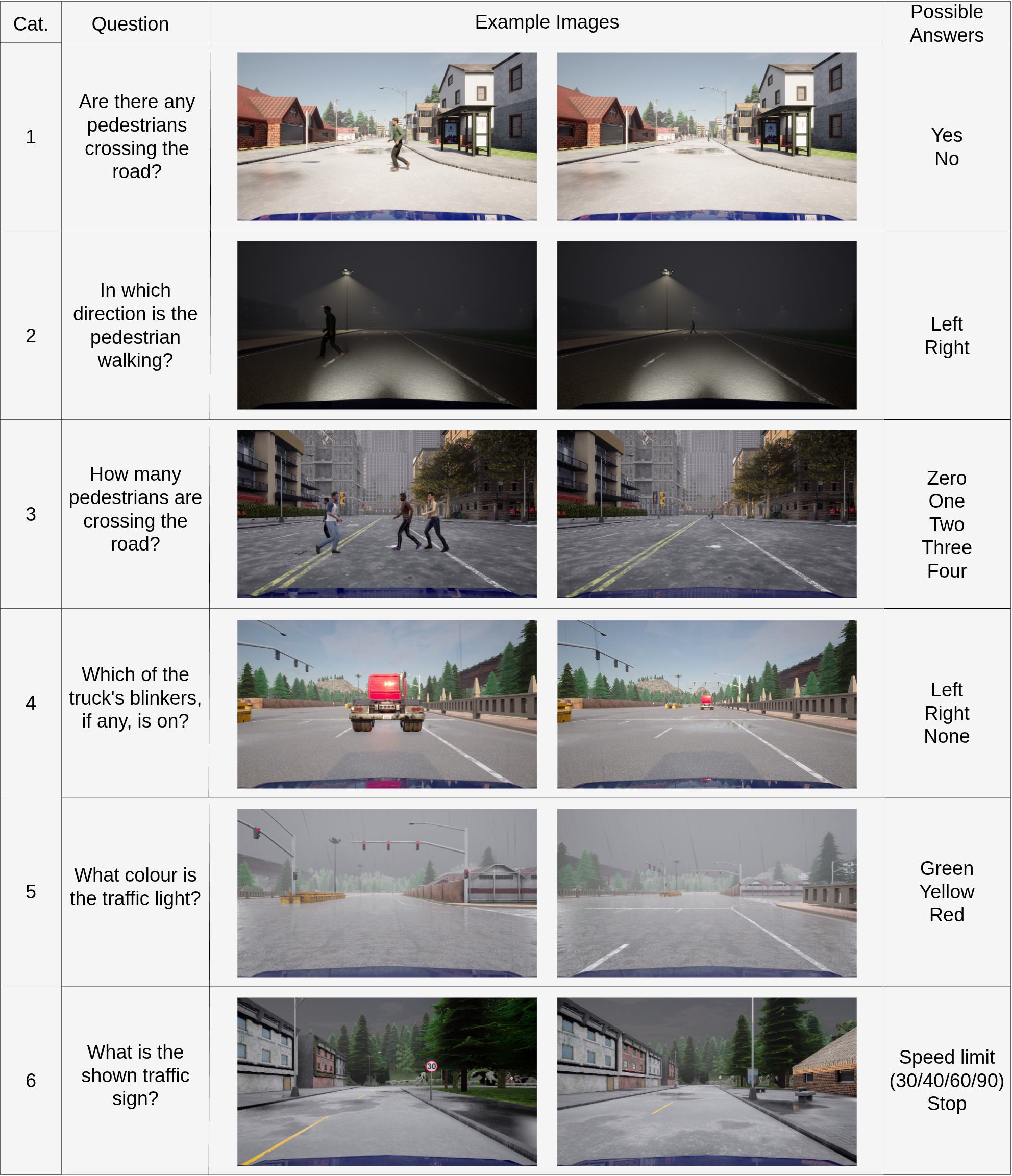}}
    \caption{The six different categories of samples in DTP-Synthetic. \label{DTP_synthetic_samples}}
\end{figure}

\begin{figure}    
\centerline{\includegraphics[width=3.5in]{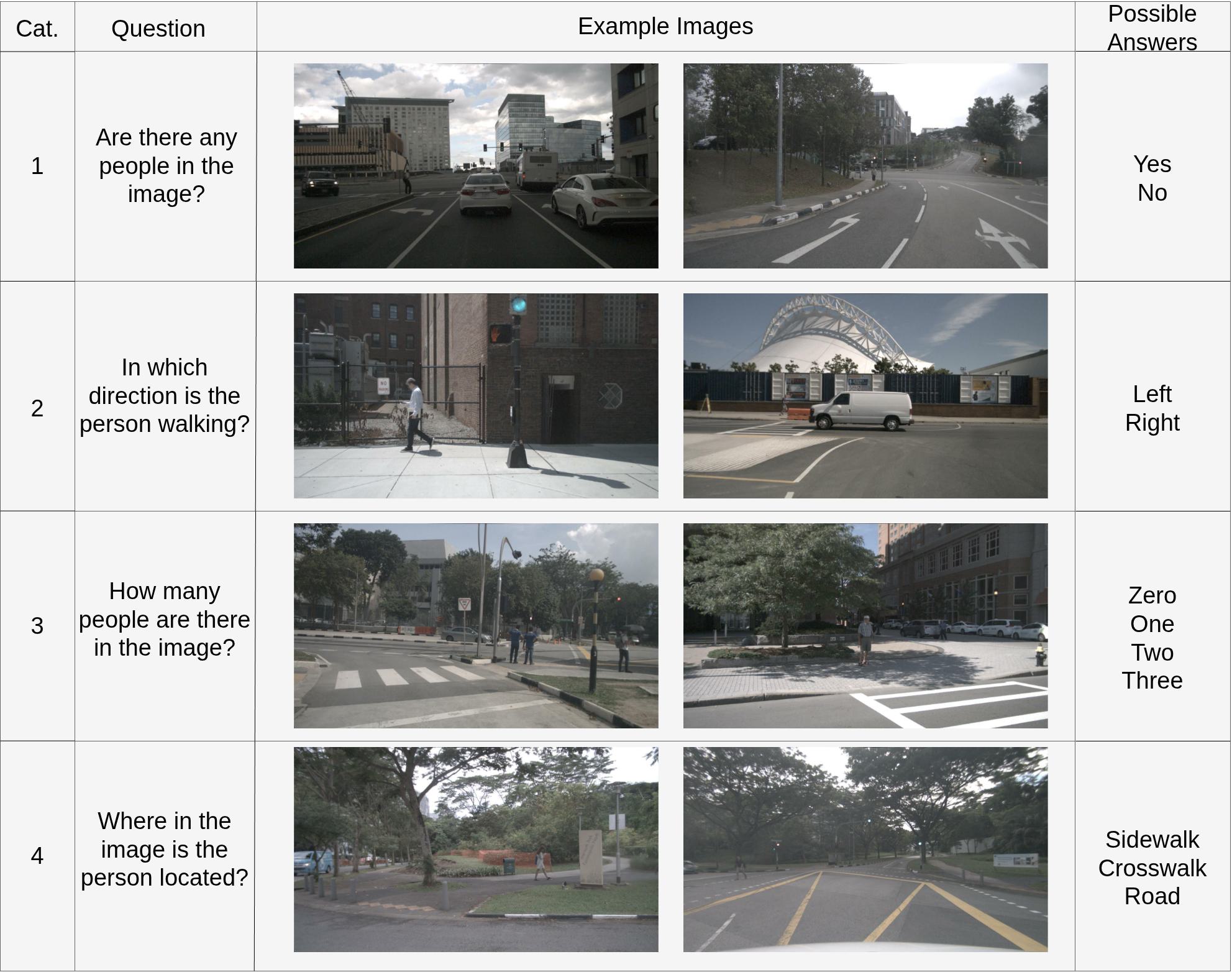}}
    \caption{The four different categories of samples in DTP-Real. \label{DTP_real_samples}}
\end{figure}

In Figures \ref{DTP_synthetic_samples} and \ref{DTP_real_samples}, we can see the ten different categories of samples in DTP-Synthetic and DTP-Real. These categories were selected to satisfy two main criteria: (a) the question must concern a simple visual concept, and (b) it must be relevant to driving decisions. For instance, knowing the direction of a walking pedestrian (category 2 in Figures \ref{DTP_synthetic_samples} and \ref{DTP_real_samples}) is crucial when planning upcoming driving actions. For example, whether a pedestrian crossing the road is moving left or right can determine whether the driver should yield or proceed, depending on the pedestrian’s position. We also included both coarse visual questions, such as the presence of a pedestrian in the scene (category 1 in Figures \ref{DTP_synthetic_samples} and \ref{DTP_real_samples}), and fine-grained ones, such as the position (left or right) of the active blinker on the vehicle ahead (category 4 in Figure \ref{DTP_synthetic_samples}). 

In Table \ref{dtp_detailed_stats}, we can get an overview of the number of samples per category and per distance in \gls{dtpqa}. As shown, the number of samples across different distances is more consistent in DTP-Synthetic. This is expected, as synthetic samples can be generated on demand, whereas the same flexibility is not available for DTP-Real, which relies on images from nuScenes. A notable observation is the significantly lower number of samples in Cat.5-Synth and Cat.6-Synth. These samples involve traffic lights and signs at various distances. Their reduced sample count stems from the limited availability of such elements in CARLA. Additionally, these categories contain no samples at a 5-meter distance. At this close range, traffic lights (due to their elevation) and traffic signs (positioned on the right side of the lane) are outside the forward-facing camera's viewing angle.

Another important feature of \gls{dtpqa} is that it maintains a balanced number of samples for each possible answer at every distance. This is something that we can observe in Table \ref{dtp_detailed_stats}, as the number of samples at each distance within a category is always divisible by the number of possible answers for that category. For example, as we can see in Table \ref{dtp_detailed_stats}, there are 396 samples of people on either a sidewalk, crosswalk, or the road (Cat.4-Real) at a distance of 30 meters in DTP-Real. The number of possible answers in this case is 3, which divides the 396 samples into three equal parts: 132 samples with a pedestrian on a sidewalk, 132 with a pedestrian on a crosswalk, and 132 with a pedestrian on the road, all at a distance of 30 meters. The same holds true for all categories and distances. The only exception is Cat.5-Synth at 10 meters, where there are 12 samples with the answer ``Red'' and 11 each for ``Yellow'' and ``Green''\footnote{We consider this slight discrepancy to be statistically insignificant and to have no meaningful impact on the evaluation of model performance for this category.}. This balance is crucial, as it prevents models from gaining an advantage based solely on language biases. For categories 1 and 3, this divisibility rule applies to the number of positive answers only, excluding the negative cases.

In the following subsections, we present a few additional details about DTP-Synthetic and DTP-Real to better understand their structure and contents.

\begin{table*}[t]
    \centering
    \caption{Number of samples per category per distance}
    \label{dtp_detailed_stats}
    \resizebox{\textwidth}{!}{
    \begin{tabular}{l *{7}{c}>{\columncolor{lightgray}}c *{7}{c}>{\columncolor{lightgray}}c}
        \toprule
        \multicolumn{1}{l}{} & \multicolumn{8}{c}{\textbf{DTP-Synthetic}} & \multicolumn{8}{c}{\textbf{DTP-Real}} \\
        \cmidrule(lr){2-9}
        \cmidrule(lr){10-17}
        \textbf{Distances} & \textbf{5 m} & \textbf{10 m} & \textbf{20 m} & \textbf{30 m} & \textbf{40 m} & \textbf{50 m} & \textbf{Negative} & \textbf{Total} & \textbf{5 m} & \textbf{10 m} & \textbf{20 m} & \textbf{30 m} & \textbf{40 m} & \textbf{50 m} & \textbf{Negative} & \textbf{Total} \\
        \midrule
        \textbf{Category 1} & 180 & 180 & 180 & 180 & 180 & 180 & 180 & 1260 & 31 & 303 & 495 & 660 & 491 & 401 & 200 & 2581 \\
        \midrule
        \textbf{Category 2} & 180 & 180 & 180 & 180 & 180 & 180 & - & 1080 & 156 & 580 & 334 & 198 & 102 & 62 & - & 1432 \\
        \midrule
        \textbf{Category 3} & 720 & 720 & 720 & 720 & 720 & 720 & 180 & 4500 & 45 & 651 & 1053 & 1056 & 432 & 162 & 200 & 3599 \\
        \midrule
        \textbf{Category 4} & 270 & 270 & 270 & 270 & 270 & 270 & - & 1620 & 147 & 684 & 447 & 396 & 255 & 240 & - & 2169 \\
        \midrule
        \textbf{Category 5} & - & 34 & 156 & 156 & 156 & 156 & - & 658 & - & - & - & - & - & - & - & - \\
        \midrule
        \textbf{Category 6} & - & 50 & 50 & 50 & 50 & 50 & - & 250 & - & - & - & - & - & - & - & - \\
        \midrule
        \rowcolor{lightgray}\textbf{Total} & 1350 & 1434 & 1556 & 1556 & 1556 & 1556 & 360 & \cellcolor{gray}9368 & 379 & 1280 & 865 & 2329 & 2310 & 2218 & 400 & \cellcolor{gray}9781 \\
        \bottomrule
    \end{tabular}%
    }
\end{table*}

\subsection{DTP-Synthetic}
DTP-Synthetic contains 8,288 unique images of 1920x1080 resolution created using CARLA \cite{Dosovitskiy2017} and 6 different questions, resulting in a total of 9,368 unique samples (image-question pairs).

Figure \ref{DTP_synthetic_samples} shows the six different categories of samples in DTP-Synthetic. As seen, the samples in DTP-Synthetic are grouped such that each unique scene is repeated multiple times, with the only variation being the distance of the object in question. In the case of static objects in categories 5 and 6 (traffic lights and signs, respectively), the ego-vehicle (and hence the camera mounted on it) must move to different distances from the object, resulting in slight changes in the background in these cases as well.

In Table \ref{dtp_detailed_stats}, we can see the number of samples per category for DTP-Synthetic. Category 3 has overwhelmingly more samples than the rest. This is because there are multiple samples per distance in each scene to cover all the possible answers. In contrast, categories 5 and 6 have significantly fewer samples due to constraints imposed by the simulator, i.e., there is a limited number of traffic lights and signs on the maps, and only a subset can be used to maintain balance in the benchmark (e.g., having an equal number of speed limit 30 and stop signs at all distances). In Table \ref{dtp_detailed_stats}, we can also see the distribution of object distances in DTP-Synthetic. As explained earlier, there are fewer samples at 5 and 10 meters because traffic lights and signs are positioned high up or at the edge of the road. The number of negative samples is much smaller, as these exist only for categories 1 and 3.

\subsection{DTP-Real}
DTP-Real contains 7,438 unique images at a resolution of 1600×900 (the native resolution of nuScenes images) and 4 different questions, resulting in a total of 9,781 unique samples. These samples were created automatically using the annotations provided by nuScenes. Figure \ref{DTP_real_samples} shows the 4 different categories of samples in DTP-Real.

DTP-Real differs from DTP-Synthetic in five key ways, aside from the obvious fact that it includes real images. First, by comparing Figures \ref{DTP_synthetic_samples} and \ref{DTP_real_samples}, we can see that the real part of the benchmark does not contain the exact equivalents of DTP-Synthetic scenes, as this is not possible due to missing annotations in nuScenes (e.g., traffic lights and signs). Additionally, we use the terms person/people instead of pedestrian, as the individuals present in nuScenes scenes are not always pedestrians (e.g., police officers, construction workers, etc.). Second, the traffic scenes depicted are much more cluttered and complex, being real compared to the simulated ones, which show an empty road with only the object in question. Third, the distance to the object in question is not always exact; therefore, we created six distance bins (i.e., 5m, 10m, 20m, 30m, 40m, and 50m) and assigned each sample to its nearest bin. Fourth, unlike DTP-Synthetic, DTP-Real does not include variations of the same scene across all distances, as this is not feasible with real data. Finally, there is a difference specific to category 3. In DTP-Synthetic, all pedestrians were spawned at the same distance from the vehicle with a very small perturbation, so they do not appear at the exact same point. However, for all of them, the distance from the vehicle remains nearly identical—that is, the characteristic distance of the sample. In contrast, since the samples in DTP-Real are based on real traffic scenes, we cannot expect all the people in the image (when there is more than one) to be at the same distance from the vehicle. To work around this, we define the characteristic distance of a sample as the average distance of all the people in the scene. Consequently, the variance in the distances of the people also becomes a factor influencing how difficult a sample is to answer (e.g., a sample with an average distance of 20m where one person is 5m away and another is 35m away is harder than a sample where both are at 20m), and hence we include it in the annotations of our data as well.

Despite these five differences, the main concept remains the same—i.e., each category contains multiple different samples where the object in question appears at varying distances or is completely absent (for categories 1 and 3).

In Table \ref{dtp_detailed_stats} we can see the number of samples per category in DTP-Real. The distribution is not uniform across the four categories due to the unequal availability of suitable images in nuScenes for each case. For example, category 2 requires a single pedestrian in the scene within a specific angle relative to the camera (to clearly indicate movement to the left or right), and there are fewer such images compared to category 3, which only requires multiple people in the scene. In Table \ref{dtp_detailed_stats} we can also see the number of samples per distance. As observed, samples depicting the relevant scene at very close distances (i.e., $<$7.5m) are less frequent by comparison.

\section{MODELS}
Our goal here is to evaluate the perception capabilities of \gls{sota} small \glspl{vlm} with fewer than 4 billion parameters, as these are the models that could potentially be deployed on local hardware within a vehicle due to their smaller size. We selected the top-performing models in the \textless 4B category from the Open \gls{vlm} Leaderboard \cite{duan2024vlmevalkit}. Table \ref{top_small_vlms} presents the top 10 performing small \glspl{vlm} on the Open \gls{vlm} Leaderboard as of March 2025.

BlueLM-V-3B \cite{Lu2025}, which ranks first, is not publicly available and was therefore excluded from our experiments. The remaining models are all publicly accessible and can be readily used for inference through the Transformers API \cite{wolf2020transformers}.

\begin{table*}[h!]
    \caption{Top 10 small Vision-Language Models on Open \gls{vlm} Leaderboard (March 2025)}
    \label{top_small_vlms}
    \centering
    \resizebox{\textwidth}{!}{
    \begin{tabular}{clccccc}
    \toprule
    \textbf{Rank} & \textbf{VLM} & \textbf{Param (B)} & \textbf{Language Model} & \textbf{Vision Encoder} & \textbf{Projector} & \textbf{Avg. Score} \\
    \midrule
    1  & BlueLM-V-3B \cite{Lu2025}          & 3.0  & BlueLM-3B             & SigLIP-400M           & - & 66.1 \\
    2  & Ovis2-2B \cite{Lu2024b}             & 2.46 & Qwen2.5-1.5B          & AIMv2 Large (310M)           & Linear layer + VE Table (370M) & 65.2 \\
    3  & Qwen2.5-VL-3B \cite{Bai2025}         & 3.75 & Qwen2.5-3B            & QwenViT (630M)              & MLP (37M) & 64.5 \\
    4  & SAIL-VL-2B \cite{Dong2025}           & 2.1  & Qwen2.5-1.5B          & InternViT-300M        & MLP (8.7M) & 61.0 \\
    5  & InternVL2.5-2B-MPO \cite{Chen2024j, Wang2024i}    & 2.0  & InternLM2.5-1.8B      & InternViT-300M-v2.5   & MLP (12.6M) & 60.9 \\
    6  & InternVL2.5-2B \cite{Chen2024j}        & 2.0  & InternLM2.5-1.8B      & InternViT-300M-v2.5   & MLP (12.6M) & 59.9 \\
    7  & Ovis2-1B \cite{Lu2024b}             & 1.27 & Qwen2.5-0.5B          & AIMv2 Large (310M)          & Linear layer + VE Table (327M) & 59.6 \\
    8  & Aquila-VL-2B \cite{Gu2024}         & 2.2  & Qwen2.5-1.5B          & SigLIP-400M           & MLP (4M) & 59.4 \\
    9  & DeepSeek-VL2-Tiny \cite{Wu2024c}    & 3.4  & DeepSeekMoE-3B        & SigLIP-400M           & MLP (7.5M) & 58.1 \\
    10 & Qwen2-VL-2B \cite{Wang2024}          & 2.0  & Qwen2-1.5B            & QwenViT (630M)              & MLP (34M) & 57.3 \\
    \bottomrule
    \multicolumn{7}{p{\linewidth}}{The \textit{Language Model} refers to the large language model (LLM) backbone used by each \gls{vlm}. The \textit{Vision Encoder} is the component that processes visual inputs. The \textit{Projector} maps the visual features into the \gls{llm}’s input space. The \textit{Avg. Score} represents the average score across eight benchmark tasks (normalised to 0 - 100, the higher the better): MMBench \cite{Liu2023d}, MMStar \cite{Chen2024l}, MMMU \cite{Yue2024}, MathVista \cite{Lu2024d}, OCRBench \cite{Liu2023e}, AI2D \cite{Kembhavi2016}, HallusionBench \cite{Guan2024}, and MMVet \cite{Yu2024b}.}
    \end{tabular}
    }
\end{table*}

\begin{figure}              
\centerline{\includegraphics[width=3.5in]{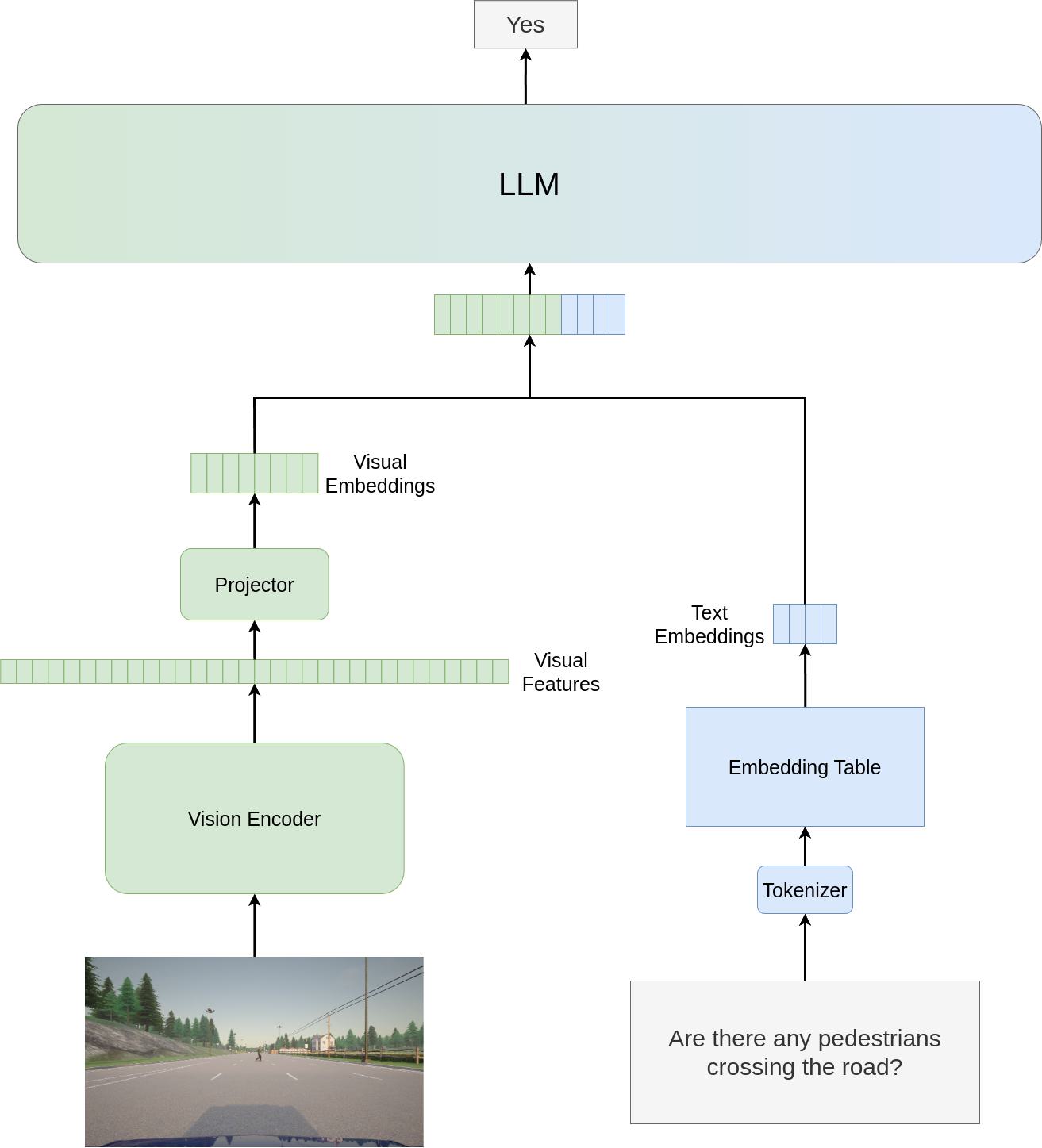}}
    \caption{The high-level architecture that most \gls{sota} \glspl{vlm} follow.\label{vlm_architecture}}
\end{figure}

All the models listed in Table \ref{top_small_vlms} follow the same high-level architecture: a Vision Encoder (typically a CLIP-like model \cite{Radford2021}) extracts visual features from the image, a projector adjusts the dimensions of those features to match the dimension of the input space of the \gls{llm}, and a pre-trained \gls{llm} processes the concatenated visual and textual inputs. This general architecture is illustrated in Figure \ref{vlm_architecture}.

For most of the models in Table \ref{top_small_vlms}, the projector is a simple MLP. The only exceptions are the two Ovis models \cite{Lu2024b}, which adopt a different approach inspired by techniques used to create text embeddings. They construct a visual embedding table (visual vocabulary × embedding dimension), then generate a token for each visual feature by computing a probability distribution over the visual vocabulary. The final visual embedding is obtained as a weighted sum of the visual vocabulary vectors, according to this distribution.

\subsection{Key Differences}
\label{subsection_key_differences}
Despite their common high-level architecture, there are some key differences in how these models process information and how they are trained. Here, we analyse these differences. It is important to note that we are excluding BlueLM-V-3B from this analysis.

The first key difference lies in how the vision encoders in these models handle images of varying resolutions. The problem arises from the fact that all vision encoders are essentially some form of Vision Transformers (ViT) \cite{Dosovitskiy2020}, and therefore, can only process images of fixed size due to the fixed size of the learned positional embeddings. There are two main strategies used by the models in Table \ref{top_small_vlms}. The first is to avoid learned positional embeddings, in which case, there is no restriction on the resolution of images that can be processed. This is the strategy followed by Qwen2.5-VL-3B and Qwen2-VL-2B, which replace learned positional embeddings with Rotary Position Embedding (RoPE) \cite{Su2024}. The second strategy, used by the remaining models in Table \ref{top_small_vlms}, is to split the image into tiles of the desired size and process them one by one, along with a thumbnail (i.e., the entire image resized to the desired resolution) for a more holistic understanding. Aquila-VL-2B \cite{Gu2024} and DeepSeek-VL2-Tiny \cite{Wu2024c} use a resolution of 384$\times$384 for each tile, whereas the rest of the models use 448$\times$448.

The second key difference, already visible in Table \ref{top_small_vlms}, is the relative sizes of the three main components (i.e., the \gls{llm}, the vision encoder, and the projector). The optimal proportions of these components remain an open research question. In most cases, however, the vast majority of parameters are allocated to the \gls{llm}. A notable exception is Ovis2-1B, which allocates less than 50\% of its parameters to its \gls{llm} backbone. As we also observe, a very small proportion of the total parameters is generally allocated to the projector, except in the case of the two Ovis models, which allocate more than 300M parameters to the projector due to their unique approach to constructing visual embeddings. To summarise, we observe the following parameter distributions:

\begin{itemize}
    \item Approximately 40\% to 90\% of the total model parameters are allocated to the \gls{llm} component (in Ovis2-1B and DeepSeek-VL2-Tiny, respectively).
    \item Approximately 10\% to 30\% is allocated to the vision encoder (in DeepSeek-VL2-Tiny and Qwen2-VL-2B, respectively).
    \item Approximately 0.2\% to 30\% is allocated to the projector (in Aquila-VL-2B and Ovis2-1B, respectively).
\end{itemize}

It is important to note here that DeepSeek-VL2-Tiny is the only model in Table \ref{top_small_vlms} that uses an \gls{llm} with \gls{moe} layers \cite{Shazeer2017}. That means that the true ratio of the computations that take place in the \gls{llm} compared to the rest of the model is much smaller than what mentioned above, as not all parameters are activated during inference for each sample.

Finally, training strategies can vary significantly, both for each component individually\footnote{The \glspl{llm} and vision encoders are typically pretrained before joint fine-tuning.} and for the entire model. However, there is a common pattern followed by almost all models in Table \ref{top_small_vlms}, which is that training is split into multiple stages. Training begins with only the projector, or both the vision encoder and projector, and uses a large amount of relatively low-quality data. Later, the \gls{llm} is included in the training process with much higher-quality data. The only exception is InternVL2.5-2B-MPO \cite{Chen2024j, Wang2024i}, which was initialized with the weights of InternVL2.5-2B \cite{Chen2024j}\footnote{In a sense, it still followed a staged pretraining process.} and then further trained on the Multimodal Preference dataset (MMPR) \cite{Wang2024i}, a high-quality, large-scale multimodal benchmark designed to enhance reasoning capabilities via chain-of-thought (CoT) prompting \cite{Wei2022}.

\section{EXPERIMENTS}
\label{section_experiments}
Here, we provide details regarding the experiments we conducted, which are divided into two categories: a) Human evaluation on \gls{dtpqa} and b) \gls{vlm} evaluation on \gls{dtpqa}.

\subsection{Human Evaluation}
Some samples in \gls{dtpqa} are difficult to answer due to the large distance of the target object, or cannot be answered at all—either because of low visibility (e.g., at nighttime, without streetlights, when a pedestrian is crossing the road 50 meters away) or due to incorrect or missing annotations in nuScenes (for DTP-Real only). Although such cases are relatively few, they lower the expected ceiling performance from 100\%, despite the trivial nature of the questions in \gls{dtpqa}. Therefore, we conducted a human evaluation experiment to estimate the maximum performance that can be expected from models on this benchmark.

There were 23 participants in this experiment, each answering 62 samples, i.e., one sample per category per distance for both the synthetic and real benchmarks, randomly sampled from \gls{dtpqa}. For each question, the possible answers were provided in multiple-choice format. No further instructions were given to the participants. It is important to note that the relatively small number of participants imposes statistical limitations. Thus, the results of the human baseline experiment should be interpreted as indicative of the achievable performance on our dataset rather than as a definitive benchmark. 

To test whether the observed differences in accuracy between humans and models were statistically significant, we performed a Monte Carlo simulation. This approach is a robust method for significance testing and is recommended when the assumptions of parametric tests may not hold, a common situation when comparing system and human performance on complex benchmarks \cite{noreen1989computer}. Specifically, we estimated how likely it would be that a model, given its overall success rate across all samples, would achieve an accuracy equal to or higher than that of human participants when evaluated on a subset of samples drawn in the same way as during the human study (i.e., one sample per distance per category for all 23 participants, allowing the same sample to be selected for multiple participants). We ran 1 million simulations per model per category and considered differences non-significant if the model matched or exceeded human accuracy on the sampled subsets more than 5\% of the time (one-sided $p > 0.05$).

\subsection{VLM Evaluation}
We tested models 2-10\footnote{As mentioned before, the weights of BlueLM-V-3B \cite{Lu2025} are not publicly available.} of Table \ref{top_small_vlms} on \gls{dtpqa}. The questions were provided to the models in multiple-choice format, and the models were explicitly prompted to respond with one of the given choices only, allowing direct comparison to the ground truth. The prompt format was as follows:

\textit{``Strictly answer with a single word only: \textless question\textgreater \ Possible answers: [\textless answers\textgreater]''}.

An exception was made for Cat.6-Synth, where the answer choices consist of more than one word (e.g., Speed limit 30). To avoid confusion, the prompt for this category was formulated as:

\textit{``\textless question\textgreater \ Choose one of the following and reply with only the answer: [\textless answers\textgreater]''}.

This prompting strategy was effective in most cases, and the models typically restricted their responses to a single word. It is also important to note that we used greedy decoding in our experiments, as the task required the models to produce a single word (or answer), making this decoding strategy the most appropriate. However, in some instances, models generated longer responses, which made exact string comparisons insufficient for evaluating correctness.

To address this, answers were categorized as follows: an answer was considered \textbf{correct} if it matched the ground truth exactly; \textbf{unclear} if the ground truth was included in the model’s response but did not match it exactly; and \textbf{incorrect} if the ground truth was not included at all in the model's response. Unclear samples were passed to an \gls{llm} (Qwen2.5-72B-Instruct \cite{Yang2024b}) along with the corresponding ground truth answers. The \gls{llm} labelled the \gls{vlm}'s answer as either ``correct'' or ``incorrect.'' We manually reviewed a subset of these results and confirmed that Qwen2.5-72B-Instruct successfully classified the unclear samples.

In a second set of experiments, we wanted to examine the effect of slight modifications to the question on the models' performance. To do this, we created three variants for each of the questions in Figures \ref{DTP_synthetic_samples} and \ref{DTP_real_samples} and reran all the experiments; the variants are shown in Table \ref{variants}. As we can see, in the first two variants we replaced only a single word with a synonym (italicised words in Table \ref{variants}). In Variant 1, we replaced the subject of the question, while in Variant 2 we replaced the verb\footnote{For Cat.1-Real, “to be” was replaced with “to appear.” This could not be achieved by changing only one word, so an additional word was added.}. In contrast, in the third variant, we rephrased the entire question while maintaining its semantic equivalence to the original.

It is important to note that this is not an extensive investigation into how question phrasing affects the performance of small \glspl{vlm} on simple visual questions. Our main goal is simply to assess whether minor modifications to the question can affect model performance at all, or whether the models are robust to this input noise. Additionally, we study modifications only to the main question itself, not to the prompt structure. We leave the study of the effect of prompt structure modifications on model performance for future work.

\begin{table*}[h!]
    \caption{Variants of the original questions}
    \label{variants}
    \centering
    \resizebox{\textwidth}{!}{
    \begin{tabular}{lccc}
    \toprule
    & \textbf{Variant 1} & \textbf{Variant 2} & \textbf{Variant 3} \\
    \midrule
    Cat.1-Synth & Are there any \textit{people} crossing the road? & Are there any pedestrians \textit{traversing} the road? & Do you see any pedestrians currently crossing the road? \\
    \midrule
    Cat.2-Synth & In which direction is the \textit{person} walking? & In which direction is the pedestrian \textit{moving}? & What direction is the pedestrian moving in? \\
    \midrule
    Cat.3-Synth & How many \textit{people} are crossing the road? & How many pedestrians are \textit{traversing} the road? & What is the number of people crossing the road?\\
    \midrule
    Cat.4-Synth & Which of the truck's \textit{indicators}, if any, is on? & Which of the truck's blinkers, if any, is \textit{activated}? & Which turn indicator on the truck is active, if any? \\
    \midrule
    Cat.5-Synth & What colour is the \textit{stoplight}? & What colour \textit{shows} the traffic light? & What is the current colour of the traffic signal? \\
    \midrule
    Cat.6-Synth & Which traffic \textit{symbol} is shown? & Which traffic sign is \textit{displayed}? & What traffic sign is visible in the image? \\
    \midrule
    Cat.1-Real & Are there any \textit{humans} in the image? & Are there any people \textit{appearing} in the image? & Do you see any humans present in the image? \\
    \midrule
    Cat.2-Real & In which direction is the \textit{human} walking? & In which direction is the person \textit{moving}? & What direction is the person moving toward? \\
    \midrule
    Cat.3-Real & How many \textit{humans} are in the image? & How many people \textit{appear} in the image? & How many people can you count in the image? \\
    \midrule
    Cat.4-Real & Where is the \textit{human} located in the image? & Where is the person \textit{positioned} in the image? & What is the person's position in the image? \\
    \bottomrule
    \end{tabular}
    }
\end{table*}

Finally, we also conducted all of the above experiments on a large \gls{sota} \gls{vlm}, namely InternVL3-78B \cite{Zhu2025d}, for comparison purposes, which at the time of this experiment ranked second on the Open \gls{vlm} Leaderboard, behind only the proprietary Gemini-2.5-Pro \cite{gemini2025}.

All experiments on small \glspl{vlm} were conducted on a single NVIDIA GeForce RTX 4070 Ti SUPER with 16GB of VRAM, with the only exception being DeepSeek-VL2-Tiny \cite{Wu2024c}, which is incompatible with FlashAttention-2 \cite{Dao2024} and therefore required more memory. For this reason, we used an NVIDIA A100-SXM4-40GB for this model. Finally, for the experiments involving InternVL3-78B, we utilised four NVIDIA H100 80GB HBM3 GPUs\footnote{Four NVIDIA A100-SXM4-40GB GPUs would likely have sufficed as well, but they were unavailable at the time due to other workloads.}.

\section{RESULTS AND ANALYSIS}
In this section, we present the results of the two experiments we conducted: (i) the evaluation of the selected models on DTPQA, and (ii) the effect of slight modifications to the questions on the models’ performance.

\subsection{Performance on \gls{dtpqa}}
\label{performance on DTPQA}
\begin{figure}              \centerline{\includegraphics[width=3.5in]{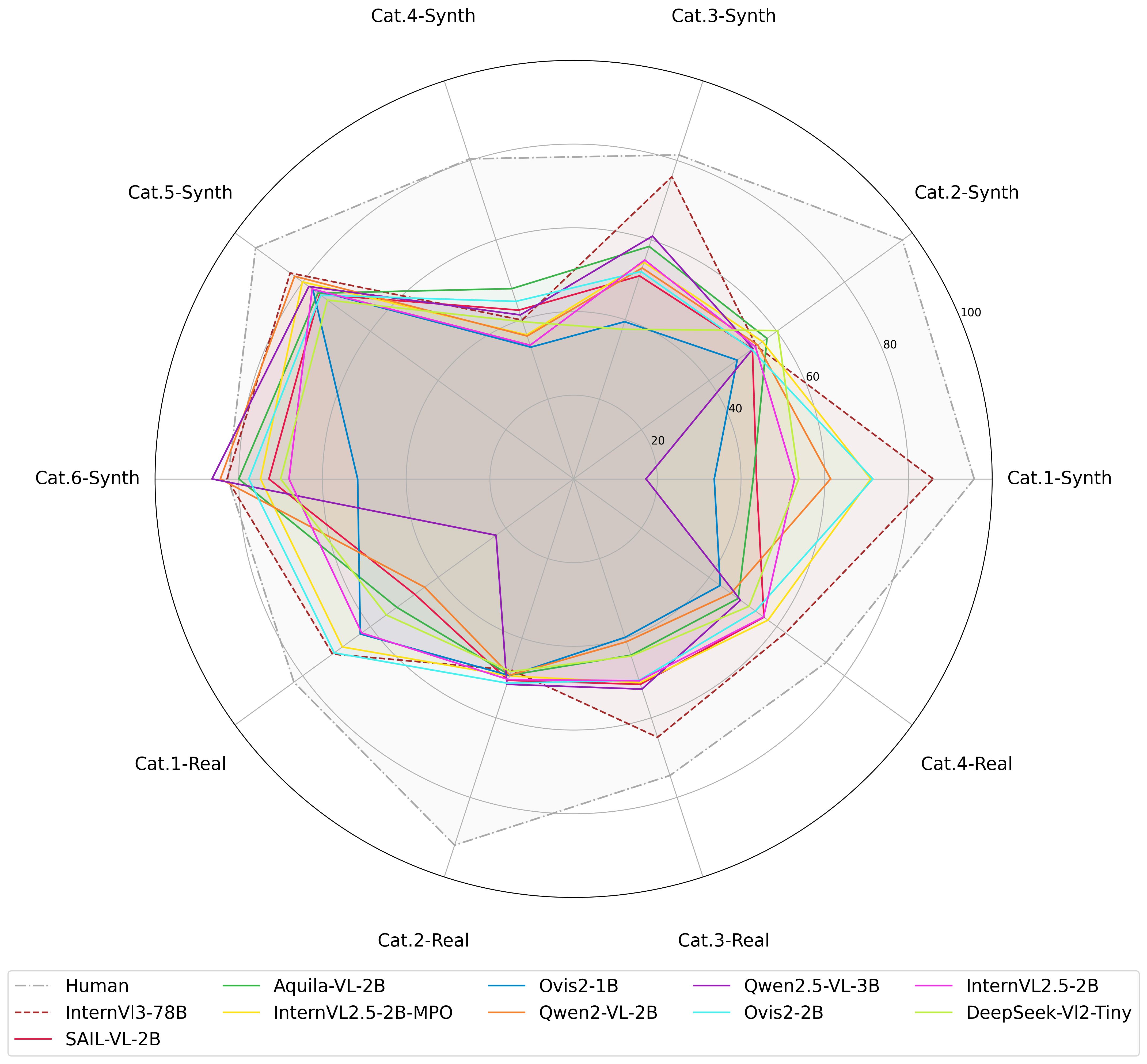}}
    \caption{Model accuracy (\%) on \gls{dtpqa}. \label{radar_chart}}
\end{figure}

Table \ref{main_results} summarises our main results, which are also visualised in Figure \ref{radar_chart}. The accuracies for each category in Table \ref{main_results} are calculated as the macro-average of the accuracies across all distances. This approach compensates for the imbalance in the number of samples across distances in some categories of DTPQA. For instance, using a micro-average would make all models appear to perform much better on Cat.1-Synth than on Cat.1-Real simply because the latter contains fewer samples at 5 meters compared to other distances.

As we can see, all models, including the large InternVL3-78B, fall short of human performance in all tasks, apart from Cat.6-Synth, where multiple models appear to reach or even slightly surpass human performance. Moreover, all differences between human and model performance were confirmed to be statistically significant with an one-sided p-value less than 0.05, as determined by the Monte Carlo simulation, except for those indicated by a dagger in Table \ref{main_results}\footnote{These exceptions occur exclusively in Cat.6-Synth, where no performance gap existed in the first place.}.

\begin{figure}              \centerline{\includegraphics[width=3.5in]{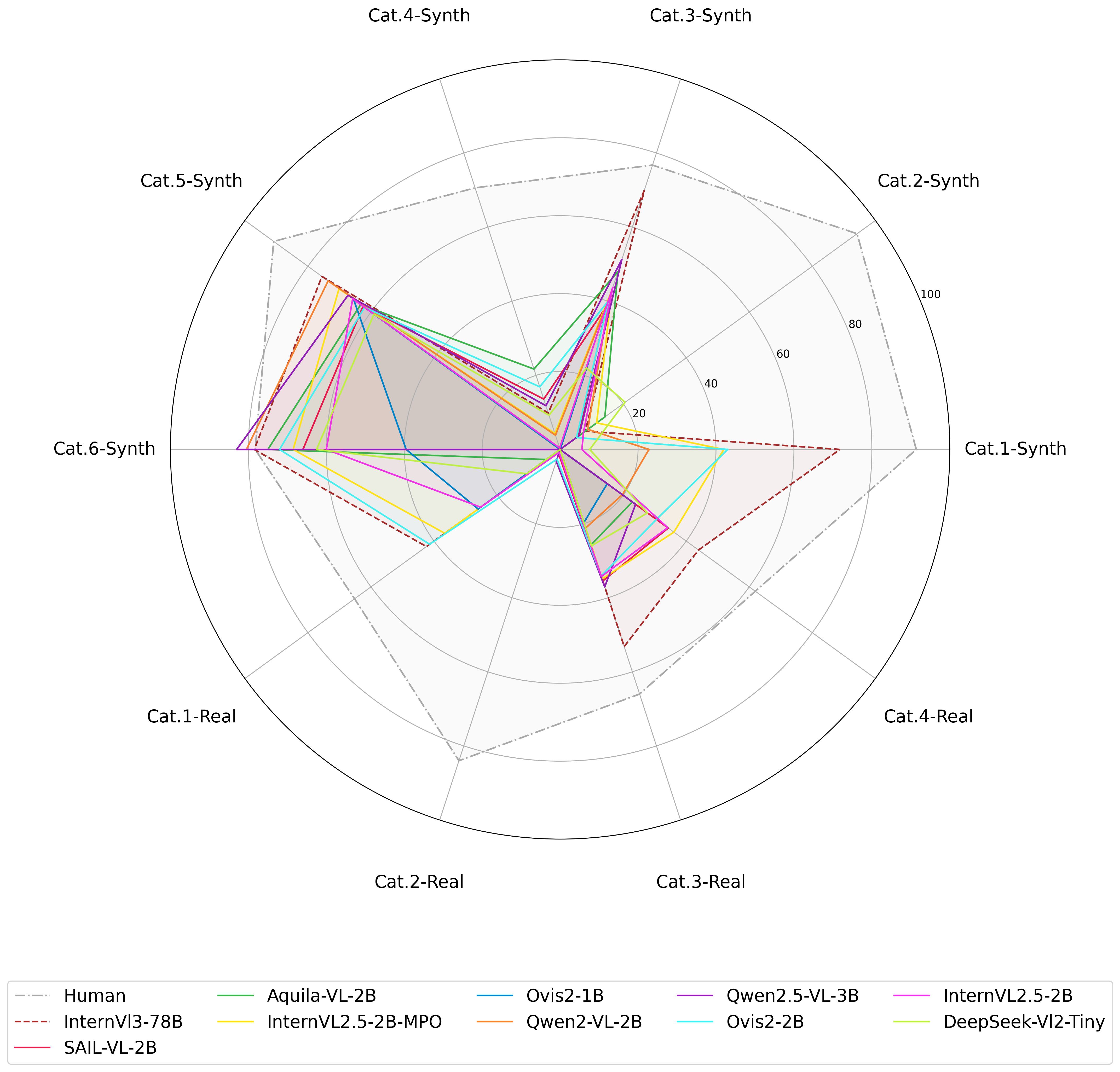}}
    \caption{Chance-corrected accuracy (\%) on \gls{dtpqa}. \label{normalized_radar_chart}}
\end{figure}

\begin{figure*}              
\centerline{\includegraphics[height=8.5in]{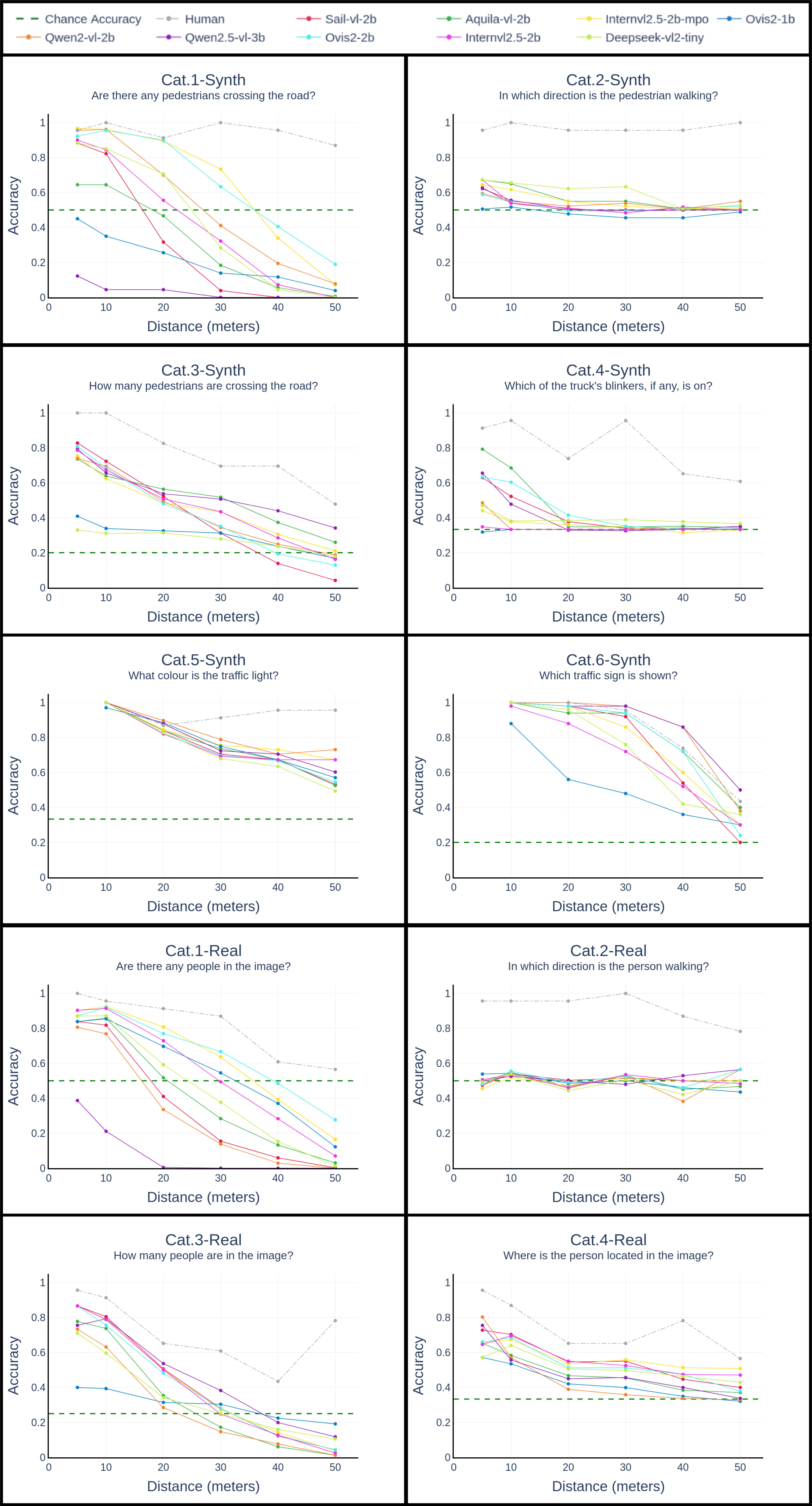}}
    \caption{Accuracy of different \glspl{vlm} across object distances for various question categories in \gls{dtpqa}. Dashed green line indicates chance level.\label{performance_by_distance}}
\end{figure*}

Figure \ref{radar_chart} does not capture the effect of the different chance accuracies for each task, and hence in Figure \ref{normalized_radar_chart} we visualise the chance-corrected accuracy of the models for each task defined as:
\begin{equation}
 a'= 
    \begin{cases}
        \frac{a_o - a_c}{1 - a_c}, & \text{if } a_o > a_c \\
        0, & \text{if } a_o \le a_c
    \end{cases}
\end{equation} 
where $a_o$ is the observed accuracy and $a_c$ is the chance accuracy. This metric reflects the proportion of samples for which the model demonstrates meaningful understanding, beyond what would be expected by random guessing. Additionally, Figure \ref{performance_by_distance} illustrates the performance of the models as a function of the distance of the object in question, highlighting again the substantial gap between small \gls{vlm} and human performance. Below we discuss the key findings based on the above results.

\begin{table*}[t] 
    \centering 
    \caption{The accuracy (\%) of small \glspl{vlm} in \gls{dtpqa}}
    \label{main_results}
    \resizebox{\textwidth}{!}{
    \begin{tabular}{l *{6}{c}>{\columncolor{lightgray}}c *{4}{c}>{\columncolor{lightgray}}c >{\columncolor{darkgray}}c}
        \toprule
        \textbf{Method} & \multicolumn{7}{c}{\textbf{DTP-Synthetic}} & \multicolumn{5}{c}{\textbf{DTP-Real}} & \textbf{Avg} \\
        \cmidrule(lr){1-1}
        \cmidrule(lr){2-8}
        \cmidrule(lr){9-13}
        \cmidrule(lr){14-14}
        & Cat. 1 & Cat. 2 & Cat. 3 & Cat. 4 & Cat. 5 & Cat. 6 & Avg & Cat. 1 & Cat. 2 & Cat. 3 & Cat. 4 & Avg & \cellcolor{white} \\
        & {\scriptsize Presence} & {\scriptsize Direction} & {\scriptsize Count} & {\scriptsize Blinker} & {\scriptsize Light color} & {\scriptsize Sign type} & \cellcolor{white} & {\scriptsize Presence} & {\scriptsize Direction} & {\scriptsize Count} & {\scriptsize Location} & \cellcolor{white} & \cellcolor{white} \\
        \midrule
        Chance     & 50.0 & 50.0 & 20.0 & 33.3 & 33.3 & 20.0 & 34.4 & 50.0 & 50.0 & 25.0 & 33.3 & 39.6 & 37.0 \\
        \midrule
        Human Performance     & 95.7 & 97.1 & 81.4 & 80.4 & 93.9 & 82.6 & 88.5 & 82.6 & 92.0 & 74.5 & 74.6 & 80.9 & 84.7 \\
        \midrule
        InternVL3-78B         & 85.9 & 54.0 & 75.9 & 39.9 & 83.7 & 82.8\textsuperscript{\dag} & 70.4 & 71.2 & 48.0 & 64.9 & 62.6 & 61.7 & 66.1 \\
        \midrule
        \multicolumn{14}{l}{\textit{Small \glspl{vlm}:}} \\
        \midrule
        Ovis2-2B             & \textbf{71.5}  & 52.6 & 52.1 & 44.6 & 74.7 & 77.6\textsuperscript{\dag} & \textbf{62.2} & \textbf{70.7} & 51.3 & 50.6 & 53.7 & \textbf{56.6} & \textbf{59.4} \\
        Qwen2.5-VL-3B        & 17.3  & 53.0 & \textbf{61.0} & 41.2 & 78.2 & \textbf{86.4}\textsuperscript{\dag} & 56.2 & 22.9 & \textbf{51.6} & \textbf{52.8} & 49.3 & 44.2 & 50.2 \\
        SAIL-VL-2B           & 43.7  & 52.8 & 51.0 & 42.4 & 75.0 & 72.8 & 56.3 & 46.9 & 50.6 & 51.6 & 56.2 & 51.3 & 53.8 \\
        InternVL2.5-2B-MPO   & 71.0  & 55.8 & 54.4 & 36.2 & 80.1 & 74.8 & 62.1 & 68.3 & 49.5 & 51.2 & \textbf{57.4} & \textbf{56.6} & 59.4 \\
        InternVL2.5-2B       & 52.8  & 53.7 & 55.0 & 33.6 & 77.2 & 68.0 & 56.7 & 62.6 & 50.4 & 50.7 & 56.1 & 54.9 & 55.8 \\
        Ovis2-1B             & 33.6  & 48.3 & 39.5 & 33.1 & 77.0 & 51.6 & 47.2 & 63.0 & 49.4 & 39.8 & 43.3 & 48.9 & 48.1 \\
        Aquila-VL-2B         & 42.9  & 57.1 & 58.4 & \textbf{47.8} & 75.5 & 80.0\textsuperscript{\dag} & 60.3 & 52.2 & 49.3 & 44.3 & 48.6 & 48.6 & 54.5 \\
        DeepSeek-VL2-Tiny    & 53.8  & \textbf{60.3} & 37.6 & 39.5 & 72.7 & 70.0 & 55.7 & 55.3 & 48.4 & 44.5 & 51.8 & 50.0 & 52.9 \\
        Qwen2-VL-2B          & 61.4  & 54.4 & 53.0 & 35.9 & \textbf{82.4} & 84.4\textsuperscript{\dag} & 61.9 & 44.0 & 49.3 & 40.9 & 46.5 & 45.2 & 53.6 \\
        \bottomrule
        \multicolumn{14}{p{\linewidth}}{Accuracies shown are macro-averaged across distances, meaning they represent the average of the accuracies within each distance. The two light grey columns show macro-averages across categories, while the dark grey column represents the macro-average of the light grey columns. \dag The difference between model performance and human performance is not statistically significant (one-sided $p > 0.05$).}
    \end{tabular}%
    }
\end{table*}

\subsubsection{Strengths and Weaknesses}
Here we summarise the main strengths and weaknesses of small \glspl{vlm} on simple visual tasks in traffic scenes, as derived from the results.\\

\noindent \textbf{Spatial Reasoning}\\
As it is clearly apparent in Figure \ref{normalized_radar_chart}, some types of questions appear to be significantly more challenging for all models compared to others. In particular, questions that require distinguishing left from right seem to be the most difficult (Cat.2-Synth, Cat.4-Synth, and Cat.2-Real), even for InternVL3-78B, which performs around chance accuracy in both Cat.2-Synth and Cat.2-Real. Especially in Cat.2-Real, all models perform close to chance, whereas human performance remains high, reaching 92.0\%.

This gap is further highlighted in the accuracy-by-distance results (Figure \ref{performance_by_distance}). As we can see, in real traffic scenes, no model appears capable of identifying the direction in which a person is walking (Cat.2-Real), even when the person is only 5 meters away. In contrast, human performance on the same task is nearly perfect up to a distance of 30 meters and only slightly declines thereafter. Similarly, no model seems able to identify which indicator is on in Cat.4-Synth when the truck is 20 meters away or more, whereas human accuracy consistently remains above 60\%, even at a distance of 50 meters. These results highlight the difficulty these models have with spatial reasoning, especially with the orientation of an object.
\\

\noindent \textbf{Traffic Light and Sign Recognition}\\
Conversely, as shown in Figure \ref{normalized_radar_chart}, most models perform very well in distinguishing colours and reading traffic signs (Cat.5-Synth and Cat.6-Synth, respectively). Based on the indicative results from the human evaluation, it appears that many small \glspl{vlm} can recognize traffic signs in the scene as effectively as humans. Additionaly, we can see in Figure \ref{performance_by_distance} that the Qwen models seem very robust at recognising traffic signs (Cat.6-Synth) up to a distance of 30 meters, as indicated by the nearly flat performance line in the corresponding category. This may reflect strong OCR capabilities in these models.

\subsubsection{Degradation of Performance with Distance}
For some question types, the performance of most models drops almost linearly with distance (Cat.1-Synth, Cat.3-Synth, Cat.5-Synth, Cat.1-Real, and Cat.3-Real). At the same time, we do not observe this pattern in human performance for Cat.1-Synth and Cat.5-Synth, where accuracy remains high even at large distances. This indicates substantial room for improving the performance of small \glspl{vlm} on these tasks at long distances. For Cat.3-Synth, Cat.1-Real, and Cat.3-Real, human performance also declines with increasing distance\footnote{An exception is the increase in accuracy from 40 to 50 meters in Cat.3-Real. We suspect this is due to noise from the relatively small number of participants in the human evaluation.}, though always maintaining a large performance gap over the best small \glspl{vlm}.

\subsubsection{Inconsistent Model Behaviour}
Looking at Qwen2.5-VL-3B in Table \ref{main_results} we notice something interesting: The model seems to have performed the worst by far on Cat.1-Synth, almost always answering ``No'' regardless of the visual input, yet achieved the best performance on Cat.3-Synth. The same pattern was observed in the corresponding categories of DTP-Real as well. Paradoxically, the model seems capable of counting multiple pedestrians in a scene but fails to detect a single one. This highlights how the behaviour of small \glspl{vlm} can sometimes be erratic and unpredictable, even when dealing with very simple visual questions. This conclusion is supported further by the results in subsection \ref{modifications_to_questions}.

\subsubsection{Other Key Findings}

\begin{itemize}
    \item The results of models that share the same vision encoder but have different \glspl{llm} (e.g., Ovis2-2B and Ovis2-1B) can differ significantly. This indicates that the perception capabilities of \glspl{vlm} depend not only on the vision encoder but also strongly on the \gls{llm} component. One would expect that, given a certain quality of visual features and the simple nature of the questions, performance should be similar even with different language models. However, this does not seem to be the case, suggesting that some crucial processing of the visual information itself occurs within the \gls{llm}.
    \item Further fine-tuning an already trained \gls{vlm} on high-quality data, without changing its architecture, can substantially improve the model’s perception capabilities as is the case for InternVL2.5-2B-MPO, which has a 3.6\% overall improvement compared to InternVL2.5-2B.
    \item Ovis2-1B and DeepSeek-VL2-Tiny seem to perform significantly worse than the other models in Cat.3-Synth. The same trend is observed for Ovis2-1B in Cat.3-Real but not for DeepSeek-VL2-Tiny. That might indicate that counting can benefit from larger language models (or more activated parameters in the \gls{llm} during inference in the case of DeepSeek-VL2-Tiny since it uses \gls{moe} layers).
\end{itemize}

\subsubsection{Performance on Negative Samples}
A noteworthy observation is that the only categories where model accuracy drops considerably below chance level are Cat.1-Synth, Cat.3-Synth, Cat.1-Real, and Cat.3-Real, hence the categories that include negative samples. At longer distances, when the model fails to detect the person or people in the image, it defaults to the answers of the negative samples (``No'' or ``Zero'' for categories 1 and 3, respectively), resulting in accuracy that drops below chance. It is therefore useful to examine the models' performance on the negative samples in these categories, which is depicted in Table \ref{negative_samples_results}.

As we can see, almost all models achieve close to perfect accuracy on the negative samples of DTP-Synthetic. Performance is slightly lower—but still very close to perfect—on the negative samples in DTP-Real. This could be due to: (a) the more cluttered and complex real-world traffic scenes, or (b) missing annotations in nuScenes, which may lead to some misclassified negative samples—i.e., samples containing a person but labelled as not containing any. The latter is most likely the reason for the lower human performance on the negative samples of Cat.1-Real and Cat.3-Real. More specifically, pedestrians at very long distances are often not annotated in nuScenes, even though they are visible if one looks carefully. Such cases most likely led to what appears as worse human performance on these samples. Nevertheless, the high accuracy of small \glspl{vlm} on negative samples is a positive sign, suggesting a low rate of visual hallucinations for all models.

\begin{table}[t] 
    \centering
    \caption{Deviation from perfect accuracy (\%) on negative samples.}
    \label{negative_samples_results}
    \setlength{\tabcolsep}{3pt}
    \begin{tabular}{lcccc}
        \toprule
        \textbf{Models} & Cat.1-Synth & Cat.3-Synth & Cat.1-Real & Cat.3-Real \\
        \midrule
        Human Performance & 0.0 & 0.0 & 13.0 & 13.0 \\
        \midrule
        Ovis2-2B & 0.0 & 0.0 & 4.0 & 1.0 \\
        \midrule
        Qwen2.5-VL-3B & 0.0 & 0.6 & 0.0 & 9.0 \\
        \midrule
        SAIL-VL-2B & 0.0 & 0.0 & 0.0 & 1.5 \\
        \midrule
        InternVL2.5-2B-MPO & 0.0 & 0.6 & 4.5 & 2.5 \\
        \midrule
        InternVL2.5-2B & 0.0 & 0.0 & 1.5 & 1.0 \\
        \midrule
        Ovis2-1B & 0.0 & 2.8 & 2.0 & 4.0 \\
        \midrule
        Aquila-VL-2B & 0.0 & 0.0 & 0.5 & 1.0 \\
        \midrule
        DeepSeek-VL2-Tiny & 0.0 & 2.2 & 0.5 & 5.0 \\
        \midrule
        Qwen2-VL-2B & 0.0 & 0.0 & 0.0 & 2.0 \\
        \bottomrule
    \end{tabular}
\end{table}

\subsubsection{Failure Cases}
\begin{figure}              
\centerline{\includegraphics[width=3.5in]{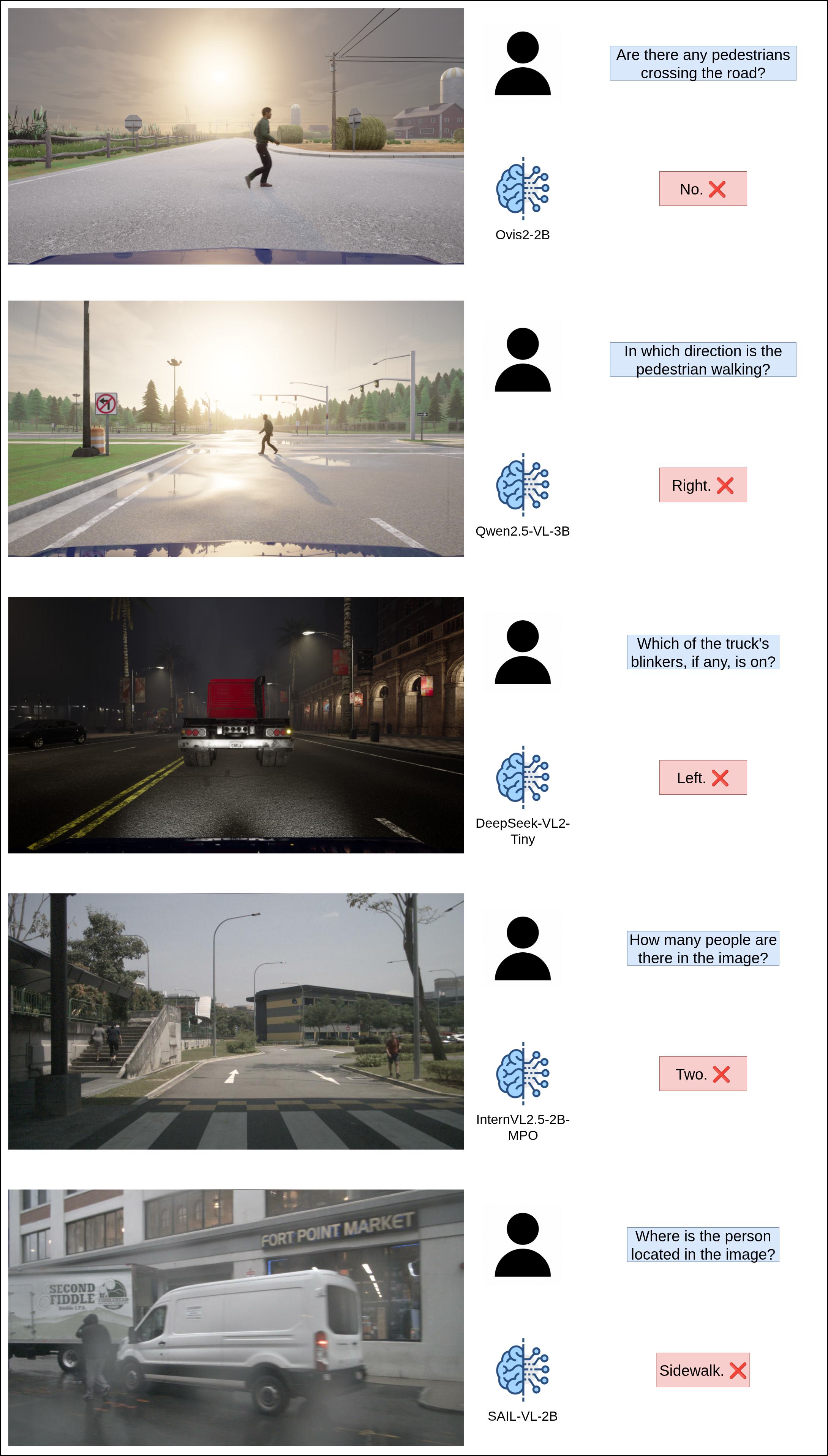}}
    \caption{Failure cases.
    \label{failure_cases}}
\end{figure}

In Figure \ref{failure_cases}, we illustrate some notable failure cases. As shown, in many instances, small \glspl{vlm} appear to be ``blind'' even when the object in question is very close ($\leq$ 20 m in all illustrated examples). These are questions that any human could always answer at a glance, but \gls{sota} small \glspl{vlm} still fail sometimes. This highlights once again that these models are not yet ready for use in automated driving. Regardless of their reasoning abilities, they cannot be trusted if they fail to consistently detect a pedestrian right in front of them (first example in Figure~\ref{failure_cases}) or determine which indicator light is on in the vehicle ahead (third example in Figure~\ref{failure_cases}).

\vspace{2em}
\noindent Overall, the above results emphasize the substantial gap between human perception and the perception systems of \gls{sota} small \glspl{vlm} in traffic scenes, and point to specific areas, such as spatial reasoning, that remain extremely challenging for these models, whereas other areas, like traffic sign recognition, match human performance.

\subsection{Impact of Question Rephrasings on Model Performance}
\label{modifications_to_questions}
Here, we present the results of the experiment where three variants of the original question were given to the models along with the visual input. The results, shown as deviations from the original accuracy, are presented in Table \ref{variants_results}. Additionally, in Figures \ref{performance_by_distance_variants_synth} and \ref{performance_by_distance_variants_real} we can see the performance by distance of the models for each variant. As mentioned earlier, this was not an extensive investigation into how question phrasing affects the performance of small \glspl{vlm} on simple visual questions; our main goal is to assess whether minor modifications to the question can affect model performance at all. Hence, all conclusions below, beyond the main one, should be seen as indicative, based on our results. Further research is definitely needed to determine why and how changes in prompts can influence the performance of small \glspl{vlm} in simple visual tasks.

\subsubsection{Primary Findings}
\begin{figure*}              
\centerline{\includegraphics[height=8.5in]{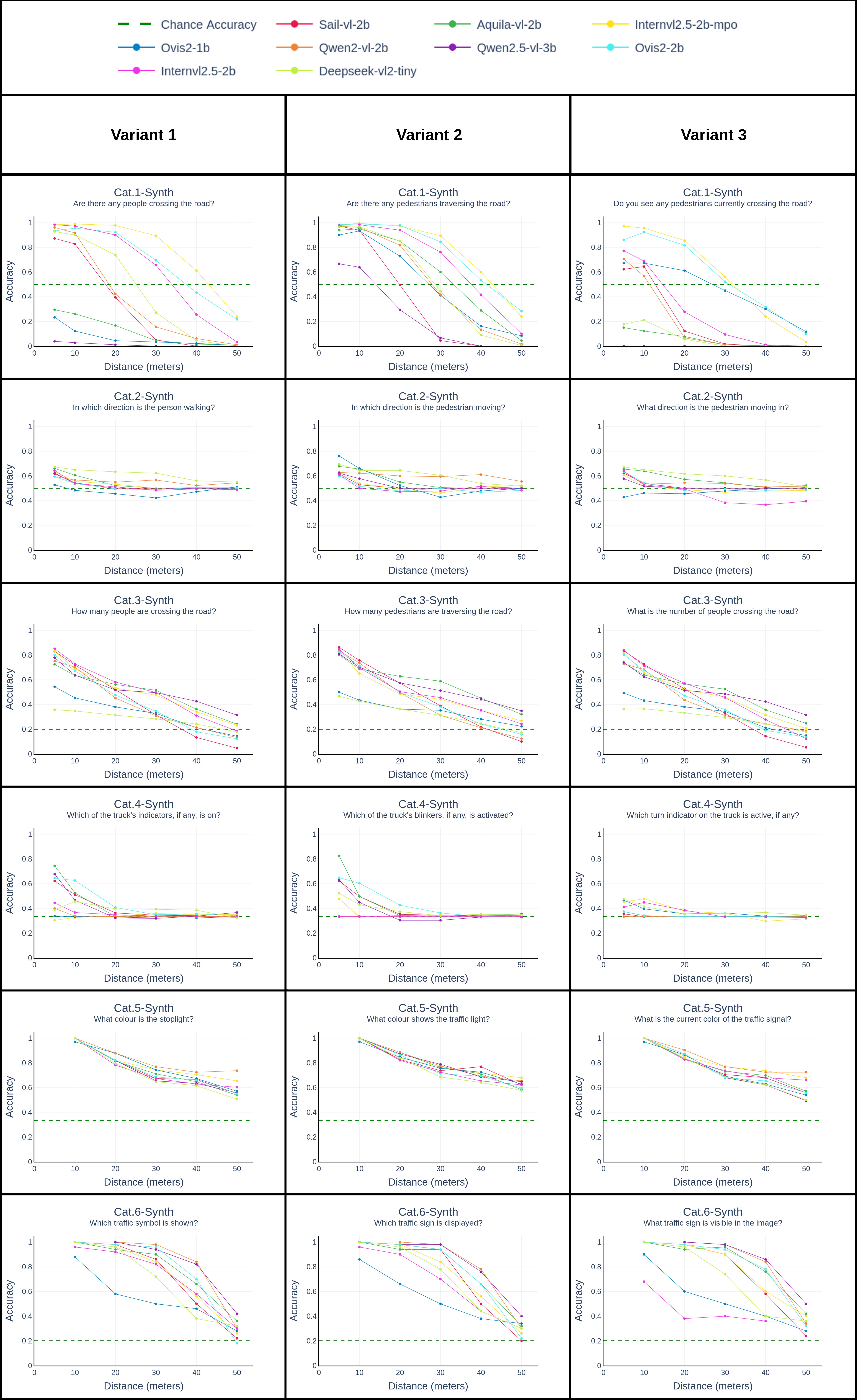}}
    \caption{Performance by distance on DTP-Synthetic for all question variations.
    \label{performance_by_distance_variants_synth}}
\end{figure*}

As we can see in Table \ref{variants_results}, changes in the question can clearly affect model performance. An extreme case is that of DeepSeek-VL2-Tiny, where simply changing the question in Cat.1-Synth from ``Are there any pedestrians crossing the road?'' to ``Do you see any pedestrians currently crossing the road?'' caused a drop of 33.2 percentage points in accuracy. However, as becomes evident in Figures \ref{performance_by_distance_variants_synth} and  \ref{performance_by_distance_variants_real}, and by comparing these figures with the original performance by distance (Figure \ref{performance_by_distance}), the overall performance patterns remain largely invariant to changes in the question. That is, models still perform much better at tasks such as identifying the colour of a traffic light or recognising a traffic sign than at determining a pedestrian’s direction or identifying which blinker is on in the vehicle ahead. We also observe the same general pattern of performance degradation with increasing distance, regardless of question phrasing. This consistency indicates that, although performance may vary across question variants, the overarching conclusions presented in subsection \ref{performance on DTPQA} remain valid.

\subsubsection{Further Observations and Patterns}
As we can see in Table \ref{variants_results}, the variance in performance caused by question variants 1 and 2 is usually smaller compared to that caused by question variant 3, though not always. This may suggest a relationship between the number of words changed and the magnitude of performance variation. Furthermore, in several cases, changing only the verb has a noticeably larger effect on performance across models than changing only the subject (e.g., Cat.1-Synth, Cat.2-Synth, Cat.3-Synth), whereas the reverse occurs only once (Cat.1-Real). This may indicate a stronger model reliance on verb choice than on other parts of speech.

Another notable finding is that the effect of changing the question is much stronger for some types of questions than others. For example, we see that changing the question for Cat.1-Synth has a significant impact on the performance of most models (for both variants), whereas it has little effect in other categories (Cat.5-Synth, Cat.2-Real, Cat.4-Real). A possible explanation for this could be that in the case of Cat.2-Real the visual information is lost at some point, so changing the prompt can't have any major impact on the performance, as this was already at chance level. Regarding Cat.5-Synth it might be the exact opposite, i.e., the visual information about the colour of the traffic light is very well encoded within the visual embeddings and hence simple changes in the prompt can't make a big difference.

We also observe that the same change in a question can produce very different effects across models. For example, the second variant in Cat.1-Synth leads to an increase in accuracy of 21.0 percentage points for Ovis2-1B but a decrease of 33.2 points for DeepSeek-VL2-Tiny, indicating that models do not respond uniformly to identical prompt changes. Nonetheless, there are interesting exceptions: the second variant of the question for both Cat.1-Synth and Cat.3-Synth consistently improves performance across all models. Notably, both apply the same modification, namely replacing ``crossing” with ``traversing”. This suggests that small \glspl{vlm} may align the visual features that encode a pedestrian crossing the road (or, in the case of Cat.3-Synth, the number of pedestrians) more strongly with the embedding of “traversing” than with that of “crossing.” Similarly, replacing “traffic light” with “stoplight” in Cat.5-Synth slightly degrades performance across all models. Finally, variant 3 of Cat.1-Synth reduces performance for almost all models (Ovis2-1B being the sole exception), often substantially. This question differs in two respects: (a) the “Do you see” format and (b) the inclusion of the word “currently.” Because variant 3 of Cat.1-Real, which also uses “Do you see” but not “currently,” mostly improves performance, the negative effect appears more likely driven by “currently,” which may bias models toward answering “No” regardless of the visual input.

The overall effect of identical changes can also differ widely between models. The average change in accuracy across all tasks ranges from 1.9 to 5.0 absolute percentage points for different models. Additionally, the results show that a much larger model size appears to mitigate but not solve this negative effect, as InternVL3-78B shows a smaller average change in accuracy of only 1.3 percentage points.

\begin{figure*}              
\centerline{\includegraphics[width=0.75\textwidth]{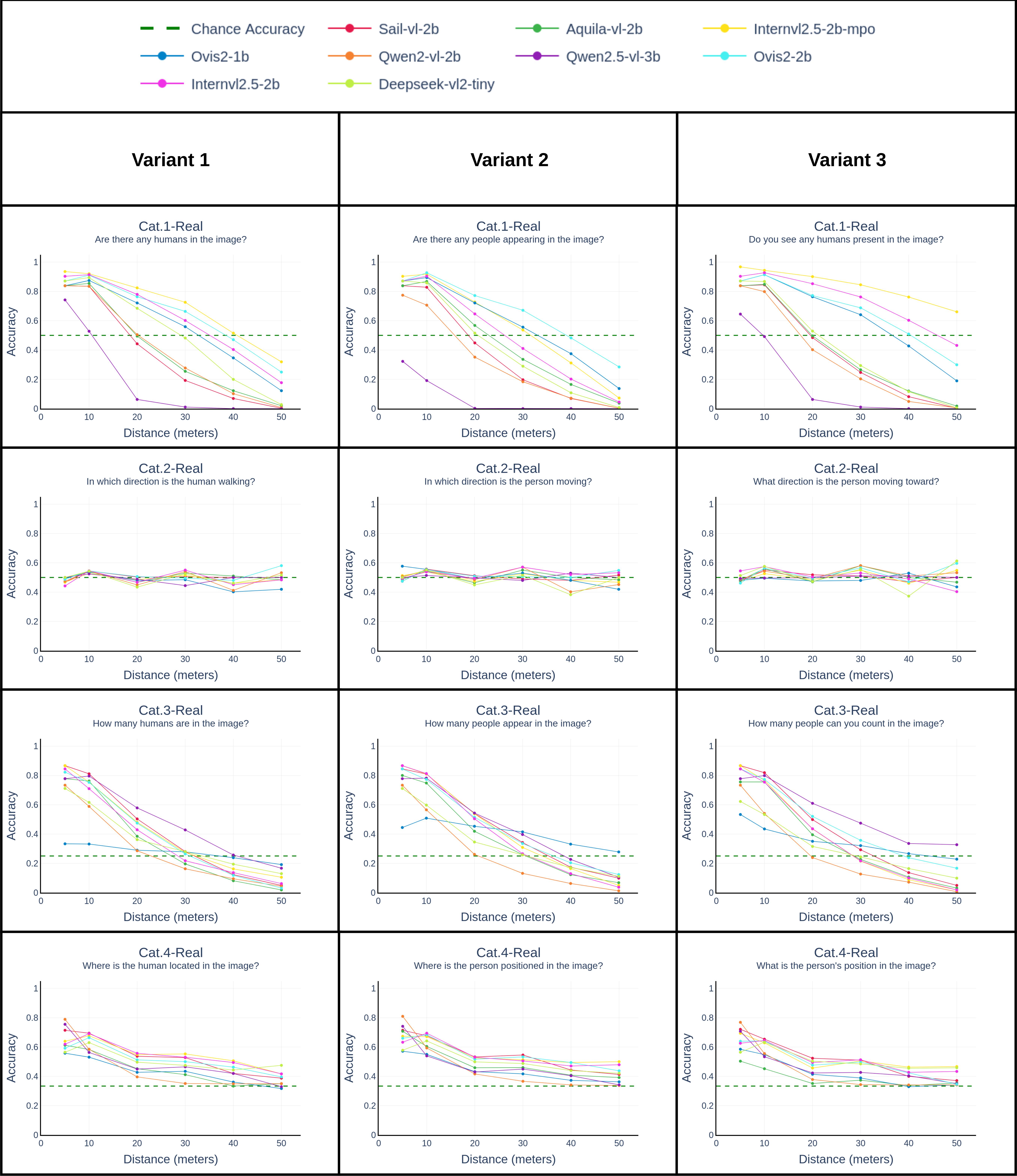}}
    \caption{Performance by distance on DTP-Real for all question variations.
    \label{performance_by_distance_variants_real}}
\end{figure*}

Finally, Table \ref{negative_samples_variants_results} presents the models’ performance, as deviation from perfect accuracy, on the negative samples of \gls{dtpqa} for all variant groups. As we can see, in most cases, performance remains close to perfect, as in the original experiment. However, there are also cases where performance drops substantially—for instance, Qwen2.5-VL-3B in Cat.3-Real when given the question variant 3. This drop helps explain the above-chance accuracy observed for this model in Cat.3-Real in Figure \ref{performance_by_distance_variants_real} (Variant 3), even at 50 meters. The model simply answered ``Zero'' much less often, increasing the chance of being correct when people were present in the scene from 1 out of 4 to 1 out of 3 (33.3\%), which matches the performance observed in Figure \ref{performance_by_distance_variants_real}.

Overall, the fact that the performance of small \glspl{vlm} can be affected, even in individual cases, by question rephrasing suggests that performance depends not only on whether the visual information required to answer the question is available (i.e., encoded in the visual embeddings), but also on how this information interacts with the text embeddings of the chosen question, even when the questions are semantically equivalent. This is clearly a concerning result for safety-critical applications, as we do not want our systems to fail simply because the task was phrased differently.

\begin{table*}[t]
    \centering 
    \caption{Deviation from original accuracy (\%) for each question variant and model.}
    \label{variants_results}
    \resizebox{\textwidth}{!}{
    \begin{tabular}{l *{30}{c} >{\columncolor{lightgray}}c}
        \toprule
        \textbf{Method} & \multicolumn{18}{c}{\textbf{DTP-Synthetic}} & \multicolumn{12}{c}{\textbf{DTP-Real}} & \textbf{Avg}\\
        \cmidrule(lr){1-1}
        \cmidrule(lr){2-19}
        \cmidrule(lr){20-31}
        & \multicolumn{3}{c}{Cat. 1} & \multicolumn{3}{c}{Cat. 2} & \multicolumn{3}{c}{Cat. 3} & \multicolumn{3}{c}{Cat. 4} & \multicolumn{3}{c}{Cat. 5} & \multicolumn{3}{c}{Cat. 6} & \multicolumn{3}{c}{Cat. 1} & \multicolumn{3}{c}{Cat. 2} & \multicolumn{3}{c}{Cat. 3} & \multicolumn{3}{c}{Cat. 4} & \cellcolor{white} \\
        & \multicolumn{3}{c}{{\scriptsize Presence}} & \multicolumn{3}{c}{{\scriptsize Direction}} & \multicolumn{3}{c}{{\scriptsize Count}} & \multicolumn{3}{c}{{\scriptsize Blinker}} & \multicolumn{3}{c}{{\scriptsize Light color}} & \multicolumn{3}{c}{{\scriptsize Sign type}} & \multicolumn{3}{c}{{\scriptsize Presence}} & \multicolumn{3}{c}{{\scriptsize Direction}} & \multicolumn{3}{c}{{\scriptsize Count}} & \multicolumn{3}{c}{{\scriptsize Location}} & \cellcolor{white} \\
        \cmidrule(lr){2-4}
        \cmidrule(lr){5-7}
        \cmidrule(lr){8-10}
        \cmidrule(lr){11-13}
        \cmidrule(lr){14-16}
        \cmidrule(lr){17-19}
        \cmidrule(lr){20-22}
        \cmidrule(lr){23-25}
        \cmidrule(lr){26-28}
        \cmidrule(lr){29-31}
        & Var.1 & Var.2 & Var.3 & Var.1 & Var.2 & Var.3 & Var.1 & Var.2 & Var.3 & Var.1 & Var.2 & Var.3 & Var.1 & Var.2 & Var.3 & Var.1 & Var.2 & Var.3 & Var.1 & Var.2 & Var.3 & Var.1 & Var.2 & Var.3 & Var.1 & Var.2 & Var.3 & Var.1 & Var.2 & Var.3 & \cellcolor{white} \\
        \midrule
        InternVL3-78B & \textcolor{darkgreen}{1.7} & \textcolor{darkgreen}{3.5} & \textcolor{red}{8.4} & \textcolor{darkgreen}{0.3} & \textcolor{darkgreen}{0.1} & \textcolor{red}{1.1} & \textcolor{darkgreen}{0.1} & \textcolor{darkgreen}{1.4} & \textcolor{red}{0.1} & \textcolor{red}{1.6} & \textcolor{red}{0.8} & \textcolor{red}{1.5} & \textcolor{red}{3.3} & \textcolor{darkgreen}{0.4} & \textcolor{red}{1.4} & \textcolor{red}{0.4} & \textcolor{darkgreen}{0.4} & \textcolor{darkgreen}{0.8} & \textcolor{darkgreen}{0.8} & \textcolor{red}{1.6} & \textcolor{red}{3.0} & \textcolor{darkgreen}{1.5} & \textcolor{darkgreen}{0.6} & \textcolor{darkgreen}{0.3} & \textcolor{orange}{0.0} & \textcolor{darkgreen}{0.1} & \textcolor{darkgreen}{1.3} & \textcolor{red}{0.7} & \textcolor{darkgreen}{0.5} & \textcolor{red}{1.0} & 1.3 \\
        \midrule
        \multicolumn{11}{l}{\textit{Small \glspl{vlm}:}} \\
        \midrule
        Ovis2-2B & \textcolor{darkgreen}{2.2} & \textcolor{darkgreen}{8.7} & \textcolor{red}{6.7} & \textcolor{red}{0.6} & \textcolor{red}{1.9} & \textcolor{red}{0.5} & \textcolor{red}{0.7} & \textcolor{darkgreen}{2.5} & \textcolor{red}{0.2} & \textcolor{darkgreen}{0.8} & \textcolor{darkgreen}{0.6} & \textcolor{red}{10.5} & \textcolor{red}{0.3} & \textcolor{darkgreen}{2.6} & \textcolor{darkgreen}{0.6} & \textcolor{red}{1.2} & \textcolor{red}{1.6} & \textcolor{darkgreen}{2.8} & \textcolor{red}{0.7} & \textcolor{darkgreen}{0.1} & \textcolor{darkgreen}{1.0} & \textcolor{darkgreen}{0.3} & \textcolor{darkgreen}{0.2} & \textcolor{darkgreen}{1.2} & \textcolor{red}{1.1} & \textcolor{darkgreen}{3.1} & \textcolor{darkgreen}{4.5} & \textcolor{red}{1.6} & \textcolor{darkgreen}{1.7} & \textcolor{red}{3.3} & 2.1 \\
        Qwen2.5-VL-3B & \textcolor{red}{1.9} & \textcolor{darkgreen}{20.8} & \textcolor{red}{3.0} & \textcolor{red}{0.5} & \textcolor{darkgreen}{0.3} & \textcolor{red}{1.4} & \textcolor{red}{1.4} & \textcolor{darkgreen}{1.5} & \textcolor{red}{2.4} & \textcolor{darkgreen}{0.3} & \textcolor{red}{2.1} & \textcolor{red}{6.9} & \textcolor{red}{4.9} & \textcolor{darkgreen}{1.7} & \textcolor{red}{4.9} & \textcolor{red}{2.8} & \textcolor{red}{4.0} & \textcolor{darkgreen}{0.4} & \textcolor{darkgreen}{10.6} & \textcolor{red}{1.2} & \textcolor{darkgreen}{8.7} & \textcolor{red}{2.3} & \textcolor{red}{1.3} & \textcolor{red}{1.5} & \textcolor{darkgreen}{3.0} & \textcolor{darkgreen}{1.0} & \textcolor{darkgreen}{1.5} & \textcolor{darkgreen}{0.4} & \textcolor{red}{0.9} & \textcolor{red}{1.9} & 3.2 \\
        SAIL-VL-2B & \textcolor{darkgreen}{1.2} & \textcolor{darkgreen}{5.6} & \textcolor{red}{9.3} & \textcolor{red}{0.1} & \textcolor{red}{0.4} & \textcolor{red}{0.1} & \textcolor{red}{0.1} & \textcolor{darkgreen}{4.7} & \textcolor{darkgreen}{0.6} & \textcolor{red}{0.6} & \textcolor{red}{1.0} & \textcolor{red}{8.7} & \textcolor{red}{0.8} & \textcolor{darkgreen}{4.1} & \textcolor{darkgreen}{0.5} & \textcolor{red}{1.6} & \textcolor{red}{0.4} & \textcolor{darkgreen}{1.2} & \textcolor{darkgreen}{1.4} & \textcolor{darkgreen}{1.4} & \textcolor{darkgreen}{3.1} & \textcolor{red}{0.3} & \textcolor{red}{0.1} & \textcolor{red}{0.2} & \textcolor{darkgreen}{0.2} & \textcolor{darkgreen}{2.4} & \textcolor{darkgreen}{0.4} & \textcolor{red}{1.5} & \textcolor{red}{0.8} & \textcolor{red}{3.2} & 1.9 \\
        InternVL2.5-2B-MPO & \textcolor{darkgreen}{10.3} & \textcolor{darkgreen}{10.2} & \textcolor{red}{5.0} & \textcolor{red}{1.9} & \textcolor{red}{3.1} & \textcolor{red}{5.2} & \textcolor{darkgreen}{4.1} & \textcolor{darkgreen}{2.8} & \textcolor{darkgreen}{2.2} & \textcolor{red}{2.4} & \textcolor{red}{0.3} & \textcolor{darkgreen}{1.8} & \textcolor{red}{1.6} & \textcolor{orange}{0.0} & \textcolor{darkgreen}{0.7} & \textcolor{red}{2.4} & \textcolor{red}{2.0} & \textcolor{darkgreen}{2.8} & \textcolor{darkgreen}{5.4} & \textcolor{red}{4.8} & \textcolor{darkgreen}{11.8} & \textcolor{red}{0.5} & \textcolor{darkgreen}{0.5} & \textcolor{darkgreen}{1.2} & \textcolor{darkgreen}{0.5} & \textcolor{darkgreen}{1.9} & \textcolor{red}{3.1} & \textcolor{red}{1.7} & \textcolor{red}{1.1} & \textcolor{red}{3.8} & 3.2 \\
        InternVL2.5-2B & \textcolor{darkgreen}{15.8} & \textcolor{darkgreen}{21.2} & \textcolor{red}{12.2} & \textcolor{red}{1.0} & \textcolor{red}{2.7} & \textcolor{red}{6.9} & \textcolor{darkgreen}{4.3} & \textcolor{darkgreen}{3.4} & \textcolor{darkgreen}{1.9} & \textcolor{darkgreen}{2.6} & \textcolor{red}{0.3} & \textcolor{darkgreen}{3.6} & \textcolor{red}{3.5} & \textcolor{red}{0.8} & \textcolor{darkgreen}{0.9} & \textcolor{darkgreen}{3.6} & \textcolor{red}{2.0} & \textcolor{red}{24.4} & \textcolor{darkgreen}{5.2} & \textcolor{red}{4.4} & \textcolor{darkgreen}{13.0} & \textcolor{red}{1.4} & \textcolor{darkgreen}{2.1} & \textcolor{darkgreen}{0.4} & \textcolor{red}{2.4} & \textcolor{darkgreen}{0.7} & \textcolor{red}{2.6} & \textcolor{red}{1.0} & \textcolor{red}{0.9} & \textcolor{red}{3.8} & 5.0 \\
        Ovis2-1B & \textcolor{red}{12.7} & \textcolor{darkgreen}{26.6} & \textcolor{darkgreen}{21.0} & \textcolor{red}{0.4} & \textcolor{darkgreen}{7.5} & \textcolor{red}{1.3} & \textcolor{darkgreen}{4.2} & \textcolor{darkgreen}{5.5} & \textcolor{darkgreen}{3.5} & \textcolor{red}{0.1} & \textcolor{darkgreen}{1.0} & \textcolor{darkgreen}{4.4} & \textcolor{red}{0.3} & \textcolor{darkgreen}{1.5} & \textcolor{red}{3.1} & \textcolor{darkgreen}{2.4} & \textcolor{darkgreen}{3.2} & \textcolor{darkgreen}{2.0} & \textcolor{darkgreen}{0.6} & \textcolor{darkgreen}{1.8} & \textcolor{darkgreen}{5.4} & \textcolor{red}{2.5} & \textcolor{darkgreen}{1.5} & \textcolor{red}{1.1} & \textcolor{red}{3.5} & \textcolor{darkgreen}{8.3} & \textcolor{darkgreen}{4.0} & \textcolor{darkgreen}{0.5} & \textcolor{darkgreen}{1.8} & \textcolor{darkgreen}{0.2} & 4.4 \\
        Aquila-VL-2B & \textcolor{red}{17.3} & \textcolor{darkgreen}{23.8} & \textcolor{red}{23.4} & \textcolor{red}{2.2} & \textcolor{red}{0.4} & \textcolor{darkgreen}{0.2} & \textcolor{red}{0.7} & \textcolor{darkgreen}{5.6} & \textcolor{red}{0.2} & \textcolor{red}{3.5} & \textcolor{red}{3.0} & \textcolor{red}{14.5} & \textcolor{red}{1.3} & \textcolor{darkgreen}{3.3} & \textcolor{darkgreen}{1.2} & \textcolor{red}{2.8} & \textcolor{red}{2.8} & \textcolor{darkgreen}{1.6} & \textcolor{red}{1.1} & \textcolor{darkgreen}{2.1} & \textcolor{red}{1.0} & \textcolor{darkgreen}{1.2} & \textcolor{darkgreen}{1.2} & \textcolor{darkgreen}{1.7} & \textcolor{darkgreen}{1.6} & \textcolor{darkgreen}{4.1} & \textcolor{darkgreen}{2.1} & \textcolor{red}{2.9} & \textcolor{darkgreen}{1.9} & \textcolor{red}{9.1} & 4.6 \\
        DeepSeek-VL2-Tiny & \textcolor{darkgreen}{1.7} & \textcolor{darkgreen}{7.9} & \textcolor{red}{33.2} & \textcolor{darkgreen}{1.2} & \textcolor{darkgreen}{0.4} & \textcolor{orange}{0.0} & \textcolor{darkgreen}{1.0} & \textcolor{darkgreen}{5.0} & \textcolor{darkgreen}{1.9} & \textcolor{red}{0.5} & \textcolor{red}{0.1} & \textcolor{red}{0.9} & \textcolor{red}{1.4} & \textcolor{darkgreen}{2.2} & \textcolor{darkgreen}{0.2} & \textcolor{red}{2.4} & \textcolor{red}{0.4} & \textcolor{red}{0.8} & \textcolor{darkgreen}{3.8} & \textcolor{red}{3.2} & \textcolor{red}{2.7} & \textcolor{darkgreen}{1.3} & \textcolor{red}{0.1} & \textcolor{darkgreen}{3.3} & \textcolor{darkgreen}{1.7} & \textcolor{darkgreen}{0.3} & \textcolor{red}{2.8} & \textcolor{red}{0.4} & \textcolor{red}{0.7} & \textcolor{red}{0.1} & 2.7 \\
        Qwen2-VL-2B & \textcolor{red}{11.0} & \textcolor{darkgreen}{0.3} & \textcolor{red}{27.8} & \textcolor{darkgreen}{1.3} & \textcolor{darkgreen}{5.8} & \textcolor{darkgreen}{0.2} & \textcolor{red}{2.1} & \textcolor{darkgreen}{0.1} & \textcolor{red}{1.8} & \textcolor{red}{1.4} & \textcolor{red}{2.5} & \textcolor{red}{2.6} & \textcolor{red}{0.2} & \textcolor{red}{2.3} & \textcolor{orange}{0.0} & \textcolor{red}{2.0} & \textcolor{red}{3.2} & \textcolor{red}{1.2} & \textcolor{darkgreen}{6.9} & \textcolor{darkgreen}{0.1} & \textcolor{darkgreen}{3.1} & \textcolor{red}{0.1} & \textcolor{darkgreen}{0.1} & \textcolor{darkgreen}{3.2} & \textcolor{darkgreen}{0.4} & \textcolor{red}{1.5} & \textcolor{red}{2.3} & \textcolor{darkgreen}{0.5} & \textcolor{darkgreen}{1.2} & \textcolor{red}{1.0} & 2.9 \\
        \midrule
        \rowcolor{lightgray}Average & 8.2 & 13.9 & 15.7 & 1.0 & 2.5 & 1.8 & 2.1 & 3.5 & 1.6 & 1.4 & 1.2 & 6.0 & 1.6 & 2.1 & 1.3 & 2.4 & 2.2 & 4.1 & 4.0 & 2.1 & 5.5 & 1.1 & 0.8 & 1.5 & 1.6 & 2.6 & 2.6 & 1.2 & 1.2 & 2.9 \\
        \bottomrule
        \multicolumn{22}{l}{Green indicates increased accuracy compared to the original performance, while red indicates decreased accuracy. The average change in accuracy for each category was calculated based only on the small \glspl{vlm}.}
    \end{tabular}%
    }
\end{table*}

\begin{table*}[t] 
    \centering
    \caption{Deviation from perfect accuracy (\%) on negative samples for question variants 1, 2, and 3.}
    \label{negative_samples_variants_results}
    \setlength{\tabcolsep}{3pt}
    \resizebox{\textwidth}{!}{
    \begin{tabular}{lcccccccccccc}
        \toprule
        & \multicolumn{4}{c}{Var.1} & \multicolumn{4}{c}{Var.2} & \multicolumn{4}{c}{Var.3} \\
        \cmidrule(lr){2-5}
        \cmidrule(lr){6-9}
        \cmidrule(lr){10-13}
        \textbf{Models} & Cat.1-Synth & Cat.3-Synth & Cat.1-Real & Cat.3-Real & Cat.1-Synth & Cat.3-Synth & Cat.1-Real & Cat.3-Real & Cat.1-Synth & Cat.3-Synth & Cat.1-Real & Cat.3-Real \\
        \midrule
        Ovis2-2B & 0.0 & 0.0 & 3.0 & 0.0 & 0.0 & 0.0 & 5.0 & 3.5 & 0.0 & 0.0 & 3.5 & 4.0 \\
        \midrule
        Qwen2.5-VL-3B & 0.0 & 0.0 & 0.0 & 9.5 & 0.0 & 1.1 & 0.0 & 6.5 & 0.0 & 0.6 & 0.0 & 52.0 \\
        \midrule
        SAIL-VL-2B & 0.0 & 0.0 & 0.0 & 1.0 & 0.0 & 0.0 & 0.0 & 3.0 & 0.0 & 0.0 & 0.0 & 2.5 \\
        \midrule
        InternVL2.5-2B-MPO & 0.0 & 0.6 & 8.5 & 2.5 & 0.0 & 1.7 & 2.0 & 2.0 & 0.0 & 1.1 & 47.5 & 2.0 \\
        \midrule
        InternVL2.5-2B & 0.0 & 0.0 & 3.5 & 1.5 & 0.0 & 1.1 & 0.5 & 1.5 & 0.0 & 0.0 & 18.5 & 0.5 \\
        \midrule
        Ovis2-1B & 0.0 & 0.0 & 1.0 & 12.5 & 0.0 & 0.6 & 2.0 & 6.5 & 0.0 & 0.0 & 2.0 & 6.5 \\
        \midrule
        Aquila-VL-2B & 0.0 & 0.0 & 0.5 & 1.0 & 0.0 & 0.0 & 1.0 & 3.0 & 0.0 & 0.0 & 0.5 & 2.5 \\
        \midrule
        DeepSeek-VL2-Tiny & 0.0 & 2.2 & 2.0 & 6.0 & 0.0 & 0.6 & 0.0 & 5.5 & 0.0 & 1.1 & 0.0 & 6.0 \\
        \midrule
        Qwen2-VL-2B & 0.0 & 0.0 & 0.0 & 2.0 & 0.0 & 0.0 & 0.0 & 1.0 & 0.0 & 0.0 & 0.0 & 1.5 \\
        \bottomrule
    \end{tabular}
    }
\end{table*}

\section{LIMITATIONS AND FUTURE SCOPE}
In this work, we quantitatively measured the strength of the perception systems of small \gls{sota} \glspl{vlm} in traffic scenes and identified specific cases where these models are particularly weak. However, we did not address the question of \textbf{why} these models fail in these particular cases or how they could be improved. The wide range of components that influence a \gls{vlm}’s perception capabilities (LLM backbone, vision encoder, projector, training strategies, training datasets, etc.) makes it extremely difficult to determine which factors are responsible for strong or weak performance. Future work could involve a mechanistic interpretability approach, where we closely examine how visual information is processed within each component of the model for each of these tasks and identify where it is ``lost'' in failure cases. Specifically, techniques such as attention visualization, probing experiments, and activation patching could be used to determine how visual information flows through the model. Based on the insights that will be gained from this analysis, we could then attempt to enhance the perception capabilities of small \glspl{vlm} in traffic scenes without compromising their performance on other tasks. Additionally, another promising direction for future work is a more comprehensive investigation of the prompt sensitivity of these models in simple visual tasks. This could involve studying a broader range of question variations as well as exploring different prompt structures—an aspect that was not examined in this study.

\section{CONCLUSION}
In this paper, we evaluate the perception capabilities of \gls{sota} small \glspl{vlm} in traffic scenes as a function of the distance of the object in question. To do so, we introduce \gls{dtpqa}, a novel \gls{vqa} benchmark with both synthetic and real images. Our results show that the perception capabilities of \gls{sota} small \glspl{vlm} remain very weak, especially in specific areas such as distinguishing left from right, and are far from being trustworthy in safety-critical applications. Additionally, we show that small \glspl{vlm} tend to be very shortsighted, as their perception capabilities degrade further with increasing distance, even in cases where human perception does not exhibit the same pattern, making them even less suitable for automated driving, where many critical objects are distant. Finally, we demonstrate that simply rephrasing a question can lead to significant performance changes in specific cases. However, the overall performance patterns remain consistent, confirming that some visual concepts are inherently more difficult for these models to grasp, regardless of how the question is phrased. Our results reveal specific weaknesses of modern small \glspl{vlm} in traffic scenes; however, further research is needed to determine the exact causes of these failures and ultimately to improve the perception capabilities of these models, which could be valuable for deployment in hardware-constrained applications.

\section{ETHICS STATEMENT}
The informed consent required for participation in this study was duly obtained from all subjects involved in the data collection process. This research adheres to the principles set forth in the Declaration of Helsinki and received ethical approval from the University of Limerick's Ethics Committee (protocol/approval number: 2025\_06\_04\_S\&E).
 
The research team acknowledges the participants' right to privacy and protection from undue exposure regarding their personal lives. To safeguard these rights, all participants were fully briefed on the objectives of the study. Furthermore, the anonymity of each participant's data has been rigorously maintained.

\newpage
\bibliography{library}

\begin{IEEEbiography}[{\includegraphics[width=1in,height=1.25in,clip,keepaspectratio]{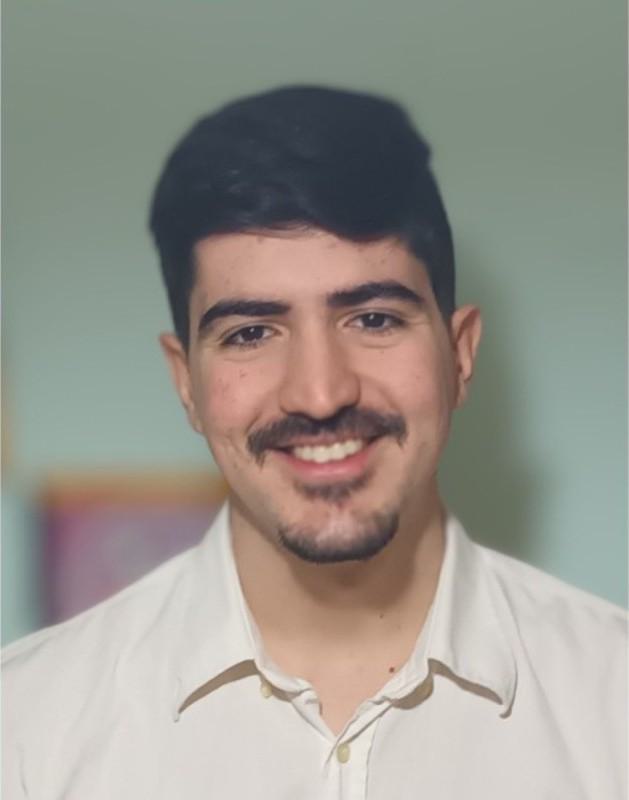}}]{NIKOS THEODORIDIS } (Graduate Student Member, IEEE) received the B.Sc. degree in physics from the Aristotle University of Thessaloniki, Thessaloniki, Greece, in 2023, and the M.Sc. degree in artificiall intelligence and machine learning from the University of Limerick, Limerick, Ireland, in 2024. He is currently working toward the full-time Ph.D. degree with the Department of Electronic and Computer Engineering, University of Limerick, Limerick, Ireland. His research interests include vision-language models and their application in automated driving.
\end{IEEEbiography}

\begin{IEEEbiography}[{\includegraphics[width=1in,height=1.25in,clip,keepaspectratio]{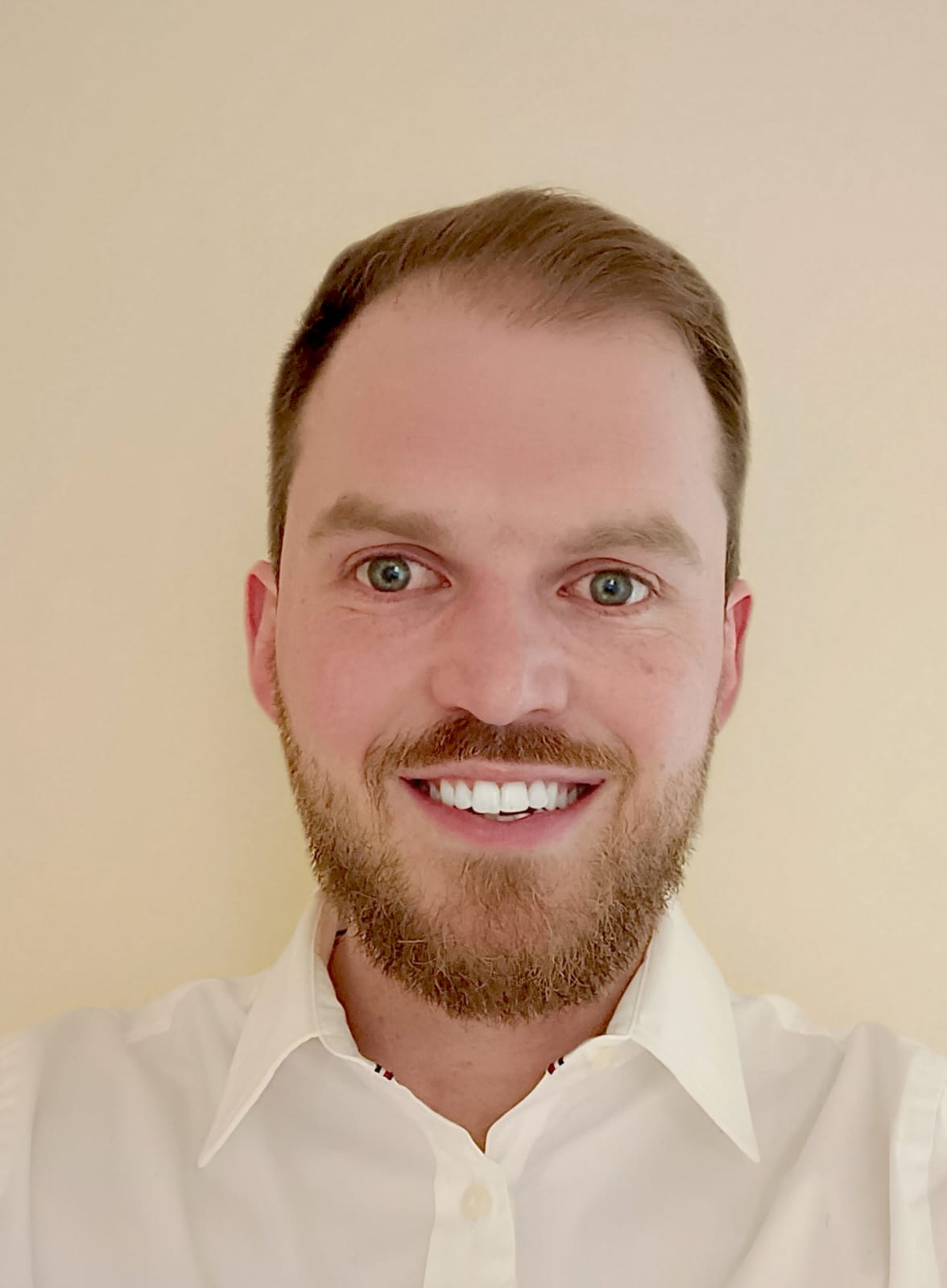}}]{TIM BROPHY } (Member, IEEE) received the
B.Eng. (Hons.) degree from the University of Gal-
way in 2018. From 2018 to 2025, he was a member
of the Connaught Automotive Research (CAR)
Group at the University of Galway, where he com-
pleted his PhD studies. In 2025, he joined the Uni-
versity of Limerick as a Postdoctoral Researcher.
His research interests include sensor availability,
computer vision, and artificial intelligence in the
context of automated vehicles.
\end{IEEEbiography}

\begin{IEEEbiography}[{\includegraphics[width=1in,height=1.25in,clip,keepaspectratio]{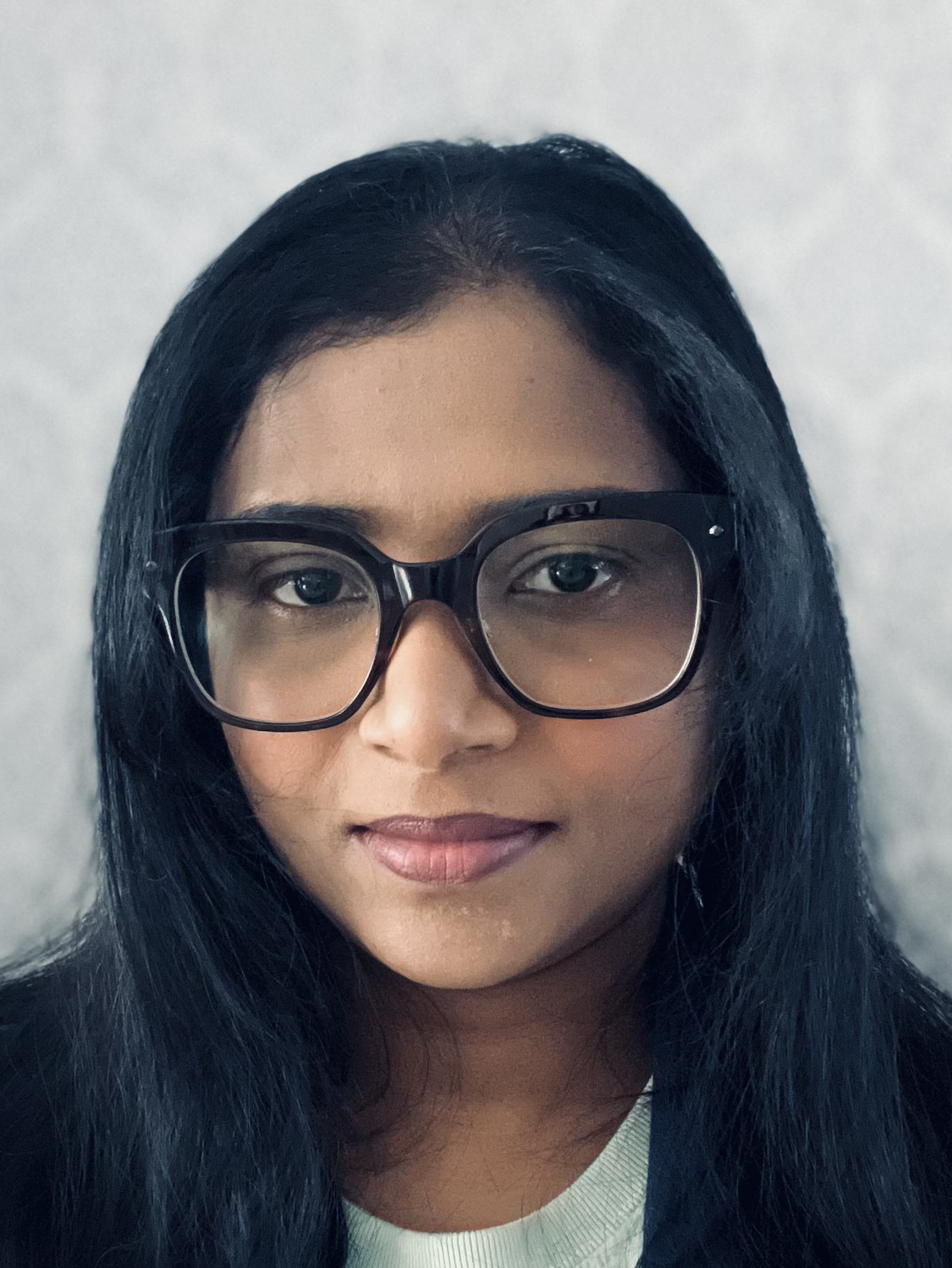}}]{REENU MOHANDAS } (Member, IEEE) received
the M.Tech. degree in digital image computing
from the University of Kerala, India, in 2014, the
M.Sc. degree in artificial intelligence from CIT,
currently Munster Technological University, Cork,
Ireland, in 2019, and Ph.D. from the Department
of Electronic and Computer Engineering (ECE),
University of Limerick, Ireland, and currently a
Postdoctoral Researcher there. Her research inter-
ests include computer vision, deep learning, in-
cremental learning, VQA models, and quantisation
and edge deployment of deep learning algorithms.
\end{IEEEbiography}

\begin{IEEEbiography}[{\includegraphics[width=1in,height=1.25in,clip,keepaspectratio]{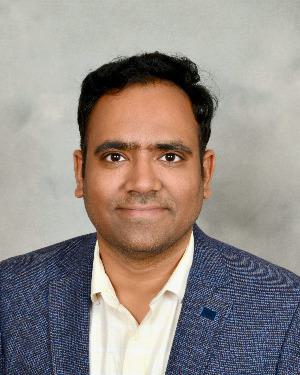}}]{GANESH SISTU } is currently a Principal AI Architect with Valeo, Tuam, Ireland, and also an
Adjunct Assistant Professor with the University of
Limerick, Limerick, Ireland. He is leading Multiple Global Research Teams working on automated
driving and parking. With more than 14 years in
computer vision and machine learning, he has authored or coauthored more than 35 publications
in top-tier conferences like ICCV and ICRA. He
plays a pivotal role in guiding the future of AI
education as an Industry Board Member for Science Foundation Ireland’s Data Science Ph.D. degree and the National M.Sc.
degree in AI programs with the University of Limerick, blending academic
insight with industry expertise.
\end{IEEEbiography}

\begin{IEEEbiography}[{\includegraphics[width=1in,height=1.25in,clip,keepaspectratio]{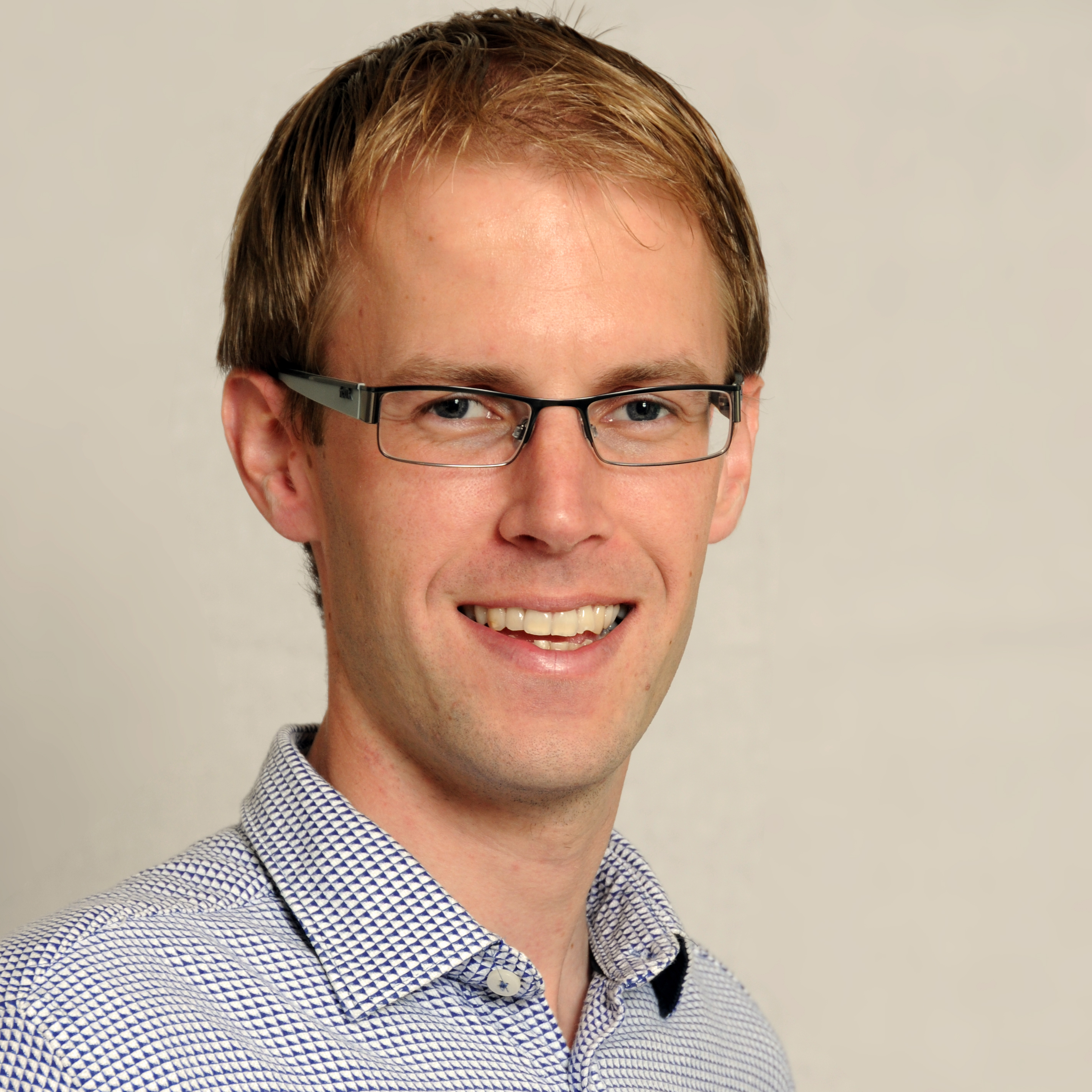}}]{Fiachra Collins } received his B.E. degree in Mechanical Engineering from University College Dublin in 2007 and PhD degree from Dublin City University in 2011. Since 2019, he has been the Geometric Perception Team Manager in Valeo Vision Systems, overseeing the software development for camera calibration and perception algorithms (3D reconstruction, localization, mapping). Prior to Valeo, from 2014-2018 he was Chief Technology Officer for a start-up company specializing in IoT and sensors, which was a spin-out of his postdoctoral research from 2011-2014. His research interests are in computer vision and data analytics. \end{IEEEbiography}

\begin{IEEEbiography}[{\includegraphics[width=1in,height=1.25in,clip,keepaspectratio]{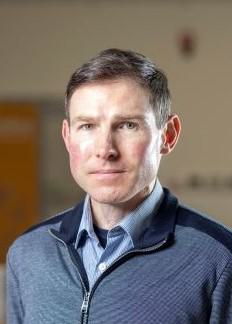}}]{ANTHONY SCANLAN } received the B.Sc. degree
in experimental physics from the National University of Ireland Galway, Galway, Ireland, in 1998,
and the M.Eng. and Ph.D. degrees in electronic
engineering from the University of Limerick,
Limerick, Ireland, in 2001 and 2005, respectively.
He is currently a Senior Research Fellow with
the Department of Electronic and Computer Engineering, University of Limerick, and has been the
Principal Investigator on several research projects
in the areas of signal processing and data converter design. His current
research interests include artificial intelligence, computer vision, and
industrial and environmental applications.
\end{IEEEbiography}

\begin{IEEEbiography}[{\includegraphics[width=1in,height=1.25in,clip,keepaspectratio]{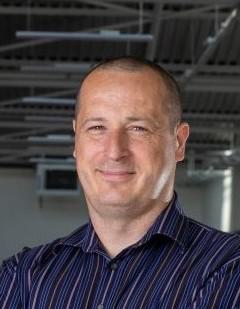}}]{CIARAN EISING } (Senior Member, IEEE) received
the B.E. degree in electronic and computer engineering and the Ph.D. degree from the NUI
Galway, Galway, Ireland, in 2003 and 2010, respectively. From 2009 to 2020, he was a Computer
Vision Architect and Senior Expert with Valeo. In
2020, he joined the University of Limerick, Limerick, Ireland, as an Associate Professor.
\end{IEEEbiography}

\end{document}